\documentclass[journal]{IEEEtran}
\usepackage{graphicx}
\usepackage{amssymb}
\usepackage{amsfonts}
\usepackage{mathrsfs}
\usepackage{amsmath}
\usepackage{algorithm}
\usepackage{algorithmic}
\usepackage{multirow}
\usepackage{booktabs}
\usepackage[table]{xcolor}
\usepackage{cite}
\usepackage{setspace}
\usepackage{bm}
\definecolor{mygray}{gray}{.9}
\definecolor{babyblueeyes}{rgb}{0.7, 0.8, 1}

\renewcommand{\algorithmicrequire}{\textbf{Input:}}

\renewcommand{\arraystretch}{1}
\ifCLASSINFOpdf
\else
\fi

\begin{document}
%
\title{Total Variation Regularized Tensor RPCA for Background Subtraction from Compressive Measurements}
%
%
%

\author{Wenfei~Cao,
        Yao~Wang,
        Jian~Sun,~\IEEEmembership{Member,~IEEE,}
        Deyu~Meng,~\IEEEmembership{Member,~IEEE,}
        Can~Yang,
        Andrzej~Cichocki,~\IEEEmembership{Fellow,~IEEE}%
        ~and~Zongben~Xu
\thanks{This work was supported in part by the Major State Basic Research Program under grant number 2013CB329404; in part by the Natural Science Foundation of China under grant numbers  11501440, 61273020, 61373114,
61472313, 61501389 and 61573270; in part by the Hong Kong Research Grant
Council under grant number 22302815, and the grant FRG2/15-16/011 from
Hong Kong Baptist University. (\emph{Corresponding author: Yao Wang.})}
\thanks{W. Cao is with the School of Mathematics and Statistics, Xi'an Jiaotong University, Xi'an 710049, China, and also with the School of Mathematics and Information Science, Shaanxi Normal University, Xi'an 710119, China (e-mail: caowenf2015@gmail.com).}
\thanks{Y. Wang is with the School of Mathematics and Statistics, Xi'an Jiaotong University, Xi'an 710049, China, and also with the Shenyang Institute of Automation, Chinese Academy of Sciences, Shenyang 110016, China (e-mail: yao.s.wang@gmail.com).}
\thanks{J. Sun, D. Meng and Z. Xu are with the School of Mathematics and Statistics, Xi'an Jiaotong University, Xi'an 710049, China (e-mail: \{jiansun, dymeng, zbxu\}@mail.xjtu.edu.cn).}
\thanks{C. Yang is with the Department of Mathematics, Hong Kong Baptist University, Kowloon, Hong Kong (e-mail: eeyang@hkbu.edu.cn).}
\thanks{A. Cichocki is with the RIKEN BSI, Wako-shi 351-0198,
Japan, and also with the Systems Research Institute, PAS, Warsaw 01-447, Poland, 
and with the Skolkovo Institute of Science and Technology (SKOLTECH),
Moscow 143025, Russia (e-mail: a.cichocki@riken.jp).}
}
\maketitle

\begin{abstract}
Background subtraction has been a fundamental and widely studied task in video analysis, with a wide range of applications in video surveillance, teleconferencing and 3D modeling. Recently, motivated by compressive imaging, background subtraction from compressive measurements (BSCM) is becoming an active research task in video surveillance. In this paper, we propose a novel tensor-based robust PCA (TenRPCA) approach for BSCM by decomposing video frames into backgrounds with spatial-temporal correlations and  foregrounds with spatio-temporal continuity in a tensor framework. In this approach, we use 3D total variation (TV) to enhance the spatio-temporal continuity of foregrounds, and Tucker decomposition to model the spatio-temporal correlations of video background. Based on this idea, we design a basic tensor RPCA model over the video frames, dubbed as the holistic TenRPCA model (H-TenRPCA). To characterize the correlations among the groups of similar 3D patches of video background, we further design a patch-group-based tensor RPCA model (PG-TenRPCA) by joint tensor Tucker decompositions of  3D patch groups for modeling the video background. Efficient algorithms using alternating direction method of multipliers (ADMM) are developed to solve the proposed models. Extensive experiments on simulated and real-world videos demonstrate the superiority of the proposed approaches over the existing state-of-the-art approaches.
\end{abstract}
\begin{IEEEkeywords}
Background subtraction, compressive imaging, video surveillance, robust principal component analysis, tensor decomposition, 3D total variation, nonlocal self-similarity.
\end{IEEEkeywords}
%
\IEEEpeerreviewmaketitle
\section{Introduction}
%
%
%
%

Since 1990s, background subtraction \cite{Piccardi2004,Bouwmans2011,Brutzer2011,Benezeth2010,Sobral2014,Bouwmans2014} has been attracting great attention in the fields of image processing and computer vision. It aims at simultaneously separating video background and extracting the moving objects from a video stream, which provides important cues for numerous applications such as moving object detection~\cite{TWang2008}, object tracking in surveillance~\cite{Beleznai2006}, etc.

\begin{figure}[htb]
\centering
\includegraphics[width=3.5in,height=1.3in]{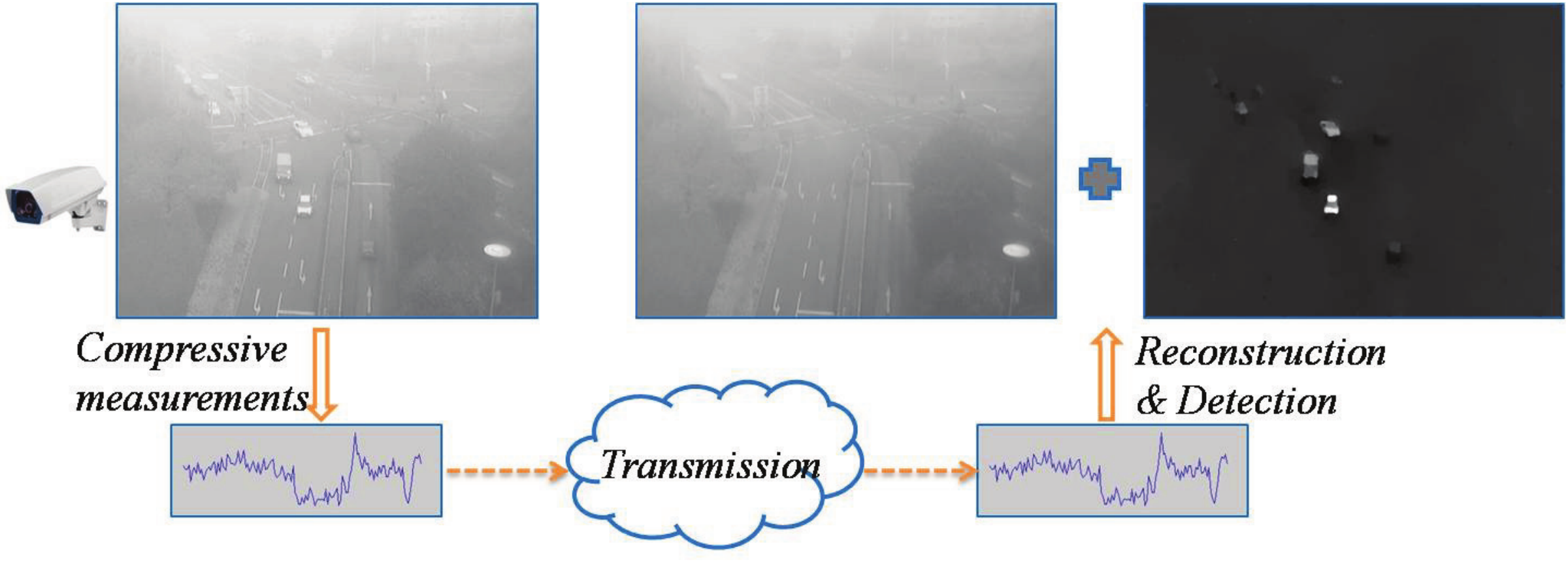}
\vspace{-3mm}
\caption{The framework of the compressive sensing surveillance system.}
\label{framework}
\end{figure}

Most of the current video background subtraction techniques consist of four steps: video acquisition, encoding, decoding, and separating the moving objects from background~\cite{Piccardi2004}. For example, Lamarre and Clark~\cite{Lamarre2002} performed background subtraction on JPEG encoded video frames using a probabilistic model; Aggarwal et al.~\cite{Aggarwal2006} considered detecting moving objects on a MPEG-compressed video using DCT coefficients of video frames.  These conventional approaches commonly implement video acquisition, coding, and background subtraction in separate procedures. This conventional scheme requires to fully sample the video frames with large storage requirements, followed by well-designed video coding and background subtraction algorithms. Recently, motivated by compressive sensing (CS)~\cite{Candes06, Donoho06, CandesTao06} in signal processing, we focus on a newly-developed compressive imaging scheme \cite{CI2006a, CI2006b, CI2006c, camera2008} for background subtraction by combining the video acquisition, coding and background subtraction into a single framework, which is called \textbf{\emph{background subtraction from compressive measurements (BSCM)}}. Figure~\ref{framework} shows an illustrative example. The video imaging system first captures compressive measurements from the scenes, and then transmits these measurements to the processing center for foreground/background reconstruction. Compared to the conventional scheme, this new scheme need not fully sense all the video voxels, and thus heavily reduces the computational and storage costs and even the energy consumption of imaging sensors.



The task of the BSCM is to reconstruct the original video with high fidelity and meanwhile accurately separate the moving objects from video background based on compressive measurements.
The objective on this task is to maximize the reconstruction and separation accuracies using as few compressive measurements as possible.
This is a heavily ill-posed inverse problem and it is necessary to discover the video prior knowledge to make this problem well-posed. There already exist some works~\cite{Cevher2008,Waters2011,Jiang2012a,Jiang2014b,Guo2014}  on the task of the BSCM.
The first seminal work was proposed by Cevher et al.~\cite{Cevher2008}, in which the dynamic adaptation of background constraint and foreground reconstruction are gracefully handled. Then, Waters et al.~\cite{Waters2011} observed that the frames in video background possess strong temporal correlation and the moving objects often occupy a small region in video foreground, and proposed a robust principal component analysis (RPCA) model to cope with this task. Guo et al.~\cite{Guo2014} further proposed an online algorithm that utilizes the spatial continuity of the supports of moving objects in video foreground. Jiang et al.~\cite{Jiang2012a,Jiang2014b} proposed a reconstruction model in which the sparsity of video foreground in the transform domain is considered. We noted that, first, all of these approaches model and characterize different video priors in a \textit{matrix} framework. Second, although these algorithms have achieved good performance, more fine video priors of background and foreground have not been fully discovered. Thus, more potential algorithms can be developed.
\begin{figure*}[htp]
\vspace{-4mm}
\centering
\includegraphics[width=6.5in,height=3.7in]{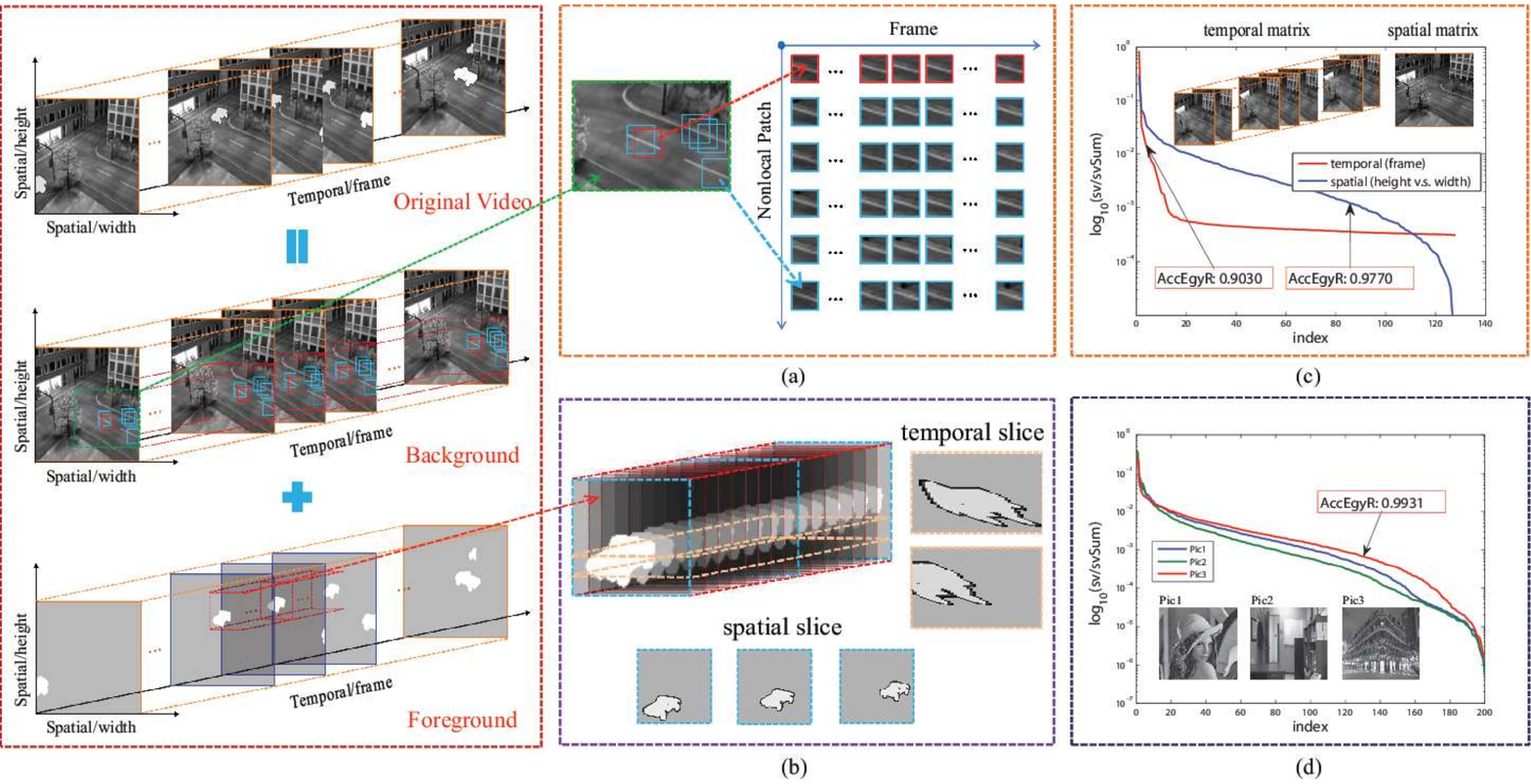}
\caption{Illustration of video priors. Please see the text for details.
(a) Non-local self-similarity prior of video background. A 3D patch has many similar 3D patches in video background. (b) The spatial-temporal continuity of video foreground. (c) Spatial-temporal correlation of video background. The red and blue curves show the singular values of two matrices, i.e., one matrix with columns of vectorized video background frames and another matrix as one frame from the video background. These two curves indicate the strong temporal correlation and moderate spatial correlation. (d) Spatial correlations in more natural images.
}
\label{mdlPri}
\end{figure*}

In this work, we propose a tensor robust principal component analysis (TenRPCA) approach  for the task of the BSCM. In this framework, we  take the video frames or video patches as the high-order tensors, and extend the robust PCA approach for matrix to the tensor-based video representation by fully investigating the domain-specific prior knowledge of surveillance videos for regularizing this inverse problem. Compared to the matrix representation of surveillance video that represents each frame as a vector, this tensor-based video representation directly takes a video frame as a matrix slice in a tensor, which preserves the spatial and temporal structure of the surveillance video.

As shown in Fig.~\ref{mdlPri}, we observed three types of priors for most surveillance videos with static backgrounds, i.e., the nonlocal similarity of 3D patches in video background, the spatio-temporal continuity of video foreground, and the spatio-temporal correlation in video background.
First, as shown in Fig.~\ref{mdlPri}(a), a 3D patch in video background possesses many similar 3D patches over the video background, and each group of similar 3D patches has strong correlation. This property is termed as~\textbf{\emph{nonlocal self-similarity of video background}}. Second, as shown in Fig.~\ref{mdlPri}(b), the moving car in video foreground is spatially continuous in both its support regions and its intensity values in these regions. Moreover, the moving car is also temporally continuous among succeeding frames. We term this prior as \textbf{\emph{spatio-temporal continuity of video foreground}}. Third, the video backgrounds are spatially and temporally correlated. In Fig.~\ref{mdlPri}(c), we show two curves (red and blue) of the singular values\footnote{For better illustration in Fig.~\ref{mdlPri}(c), we normalize the singular values of a matrix by enforcing their summation to be one.} of two matrices, i.e., one matrix with columns of vectorized video background frames and another matrix as one frame from the video background. The drastically decaying trend of the red curve indicates the strong  temporal correlation among the video background frames and the slow decaying trend of the blue curve indicates the weak correlation in the spatial domain. Let us define the accumulation energy ratio of top $k$ normalized singular values as $\text{AccEgyR}= \sum_{i} nsv_i$, where $nsv_i$ is the $i$-th normalized singular value, i.e., $nsv_i=sv_i/\sum_{i}sv_i$ and $sv_i$ is the $i$-th singular value. The arrow box for the red curve indicates that only top 3 singular values can attain the ratio 0.9030 while the arrow box for the blue curve indicates that top 84 singular values can attain the ratio 0.9770. These quantitative values justify that the video background has strong correlation among its frames and  each video background frame has weak spatial correlation. Fig.~\ref{mdlPri}(d) further exhibits
the weak correlations in other natural images. We term this observation as~\textbf{\emph{spatio-temporal correlation of video background}}.

Based on the aforementioned video priors, we model the BSCM in a tensor RPCA framework (TenRPCA) using Tucker decomposition technique. With our model, a video volume represented by a tensor is decomposed into a background layer based on its spatial-temporal correlation and foreground layer based on its spatial-temporal continuity. We design a Tucker decomposition approach to model the spatio-temporal correlation in video background and a 3D total variation (TV) term to enforce the spatio-temporal continuity of video foreground. Along this idea, we propose two TenRPCA models by representing video background as a single tensor and a few patch-level tensors over groups of similar 3D patches, which are dubbed as holistic tensor RPCA model (H-TenRPCA) and patch-group-based tensor RPCA model (PG-TenRPCA) respectively. We design efficient algorithms using the alternating direction method of multipliers (ADMM) to optimize these proposed models. The experiments on synthetic and real videos demonstrate that our proposed two TenRPCA models achieve higher reconstruction and background/foreground separation accuracies with fewer compressive measurements than the existing state-of-the-art approaches. Moreover, the PG-TenRPCA model generally works better than the H-TenRPCA model which indicates the effectiveness of our modeling of  the nonlocal self-similarity of video background.

Our \emph{contributions} can be summarized as four folds:
First, to the best of our knowledge, we are the first to model the BSCM task in a tensor robust PCA framework. Compared to the matrix-based video representation, this tensor-based video representation well preserves the spatial-temporal structures of video, which enables us to fully characterize the priors of video spatial-temporal structures in our framework. Second, we fully investigate the video priors for the BSCM task. We design a 3D total variation (TV) term to encode the spatio-temporal continuity of video foreground, and a Tucker decomposition approach to model
the spatio-temporal correlation of video background.
Third, based on the observation of nonlocal self-similarity of video background, we design a patch-level background model using joint Tucker decomposition over groups of similar 3D patches to model the strong correlations among similar 3D patches. This model significantly outperforms our holistic TenRPCA model which represents the video background as a single tensor.
Finally, based on ADMM with the adaptive scheme, we design efficient algorithms to solve the proposed models, and achieve  superior performance over the existing methods on various video data sets,
especially when the sampling ratio is very low.

The remaining of this paper is organized as follows. In Section II, the related works will be discussed.
In Section III, the general framework of the BSCM will be reviewed. Our models and their motivations will be presented in Section IV. In Section V, efficient algorithms will be designed to solve the proposed models. In Section VI, extensive experiments on various surveillance video data sets will be conducted to substantiate the superiority of the proposed models over the other existing ones. This paper will be concluded with some discussions on future work in Section VII.
\section{Related work}
\subsection{Background Subtraction without Compressive Imaging}
Various approaches for background subtraction using conventional imaging cameras have been developed since 1990s and obtained a wide range of applications in many fields. These approaches can be mainly categorized into the following five classes: the basic approach, the statistical approach,  the fuzzy approach, the neural and neuro-fuzzy approach, and the subspace learning
approach~\cite{Piccardi2004,Bouwmans2011,Brutzer2011,Benezeth2010,Sobral2014,Bouwmans2014}.

Among these traditional approaches, the subspace learning approach has been attracting wide attentions in the field of machine learning and computer vision. One classical work on this task was proposed by Oliver et al.~\cite{Oliver2000}, which uses an eigenspace (PCA) idea to model the background. Aiming at remedying the outlier and heavy noise issue, Candes et al.~\cite{candes2011} proposed robust principal component analysis (RPCA) to resist the gross sparse noise. This seminal work has triggered a tremendous interest in dealing with background subtraction using different formulations of RPCA. For example, 
the Markov random field (MRF) regularized RPCA technique was proposed in Zhou et al.~\cite{Zhou2013},
a novel block sparse RPCA formulation was proposed in~\cite{zhi2014}, total variation regularized RPCA and matrix factorization methods were respectively proposed in Cao et al.~\cite{XiaoChun2015} and Guo et al.~\cite{XiaoJie2014}, and the probabilistic versions of RPCA were proposed in Ding et al.~\cite{Ding2011} and Babacan et al.~\cite{Babacan2012}, respectively. In the recent work, Zhao et al.~\cite{Zhao2014} proposed a new probabilistic variant
by extracting multi-layer structures with certain physical meanings using the mixture of Gaussians (MOG).

To meet the real-time requirements in practical applications, various online subspace learning approaches were developed.
Rymel et al.~\cite{Rymel2004} and Li et al.~\cite{Li2006} respectively proposed an incremental PCA method to handle the newly coming video streams.
By constraining the subspace on Grassmannian manifold, Balzano et al.~\cite{Balzano2010, he2011, he2012, Balzano2014} proposed two efficient approaches named GROUSE and GRASTA respectively, to deal with  online subspace identification and tracking (SIT) task. Additionally, it was reported that the proposed GROUSE and GRASTA can effectively achieve the real-time background subtraction through sampling the voxels of video sequence. Xu et al.~\cite{Xu2013} further proposed an updated version of GRASTA by modeling the contiguous structure of supports of video foreground using group sparsity. Chi et al.~\cite{Chi2013} developed an online parallel SIT algorithm using recursive least squares technique for real-time background subtraction.
\subsection{Background Subtraction with Compressive Imaging} Recently, multiple studies have been carried out for
the background subtraction problem from the perspective of compressive imaging, in which it is required to simultaneously perform background subtraction and video reconstruction. The first seminal work was considered by Cevher et al.~\cite{Cevher2008},
in which the dynamic adaptation of background constraint and foreground reconstruction are gracefully handled.
Recently, based on the theoretical results of $\ell_1$-$\ell_1$ minimization~\cite{Mota2014}, Mota et al.~\cite{Cevher2015} proposed
an efficient adaptive-rate algorithm to deal with the BSCM task.
Additionally,  a series of work have been proposed based on the matrix RPCA technique.
Waters et al.~\cite{Waters2011} integrated the matrix RPCA methodology into the framework of the BSCM and then developed a greedy algorithm called SpaRCS to solve the resulting model. Guo et al.~\cite{Guo2014} developed an online RPCA algorithm that models the spatial continuity prior of moving objects in the foreground. Jiang et al.~\cite{Jiang2012a,Jiang2014b} proposed a new RPCA model in which the sparsity of video foreground in the transform domain is considered based on certain practical requirements.

The matrix RPCA approaches for the BSCM commonly model the video as a matrix with columns of vectorized video frames. Although the matrix RPCA methodology has been an increasingly useful technique, it fails in fully exploiting the prior knowledge on the intrinsic structures of video after vectorizing the video frames. Our proposed tensor RPCA approach considers more extensive spatio-temporal prior knowledge of video background and foreground using tensor representation of video. Such full utilization of prior information makes our approach capable of achieving a better video reconstruction quality and simultaneously detecting  the moving objects in foreground from a limited number of compressive measurements, as will be shown in Section VI. We also noted that the tensor compressive sensing models were recently proposed in~\cite{Tencs2012, Friedland2014}. But they are significantly different from our models, because these model are designed for the image/video compressive sensing task instead of the more complex BSCM task considered in this paper.
\section{The General Framework of the BSCM}
In this section, we will present the general framework for the BSCM task.
We will mainly focus on the mathematical modeling and algorithm design in the pipeline of BSCM, i.e., reconstruct video foreground and background from compressive measurements. In the followings, we will introduce the basic components of the BSCM task, including the representation of video volume, compressive operator, and video reconstruction and separation.

\subsection{Video Volume}
Video frames within a short period are collected as a video volume. If the video frame has a single channel, then the video volume can be represented as a 3-order tensor $\mathcal{X}_0:=\{\textbf{X}_{0}^{1}$,$\textbf{X}_{0}^2,...,\textbf{X}_{0}^{D}\}$, where each matrix $\textbf{X}_{0}^{i} \in \Re^{H\times W}(i=1,2,\cdots,D)$ represents $i$-th frame. $H$ and $W$ denote the height and width of a frame and $D$ denotes the number of frames.  This tensor has 3 modes
including height, width and time.  We assume that the video volume to be reconstructed can be separated into a static component (video background) $\mathcal{X}_1$, and a dynamic component (video foreground) $\mathcal{X}_2$, i.e., $\mathcal{X}_0:=\mathcal{X}_1 + \mathcal{X}_2$, where $\mathcal{X}_1:=\{\textbf{X}_{1}^{1},\textbf{X}_{1}^{2}\cdots,\textbf{X}_{1}^{D}\}$ and $\mathcal{X}_2:=\{\textbf{X}_{2}^{1},\textbf{X}_{2}^{2}\cdots,\textbf{X}_{2}^{D}\}$. In the following, we denote the vectorization of a video volume $\mathcal{X}_0$ by $\textbf{x}_0:=[\textbf{x}_{0}^{1};\textbf{x}_{0}^{2};\cdots,\textbf{x}_{0}^{D}]$, and the vectorization of video background and foreground by $\textbf{x}_1:=[\textbf{x}_{1}^{1};\textbf{x}_{1}^{2};\cdots,\textbf{x}_{1}^{D}]$ and $\textbf{x}_2:=[\textbf{x}_{2}^{1};\textbf{x}_{2}^{2};\cdots,\textbf{x}_{2}^{D}]$, respectively.

\subsection{Compressive Operator}
Compressive operator can be considered as the effective encoding of video volume.
Currently, how to design a high quality compressive operator is a crucial research topic in the CS community; see~\cite{Do2012}.
For video data, the compressive measurements $\textbf{y}$ can be obtained by
\begin{equation}
\textbf{y}=\mathcal{A}(\mathbf{x}_0),
\label{obser}
\end{equation}
where $\textbf{y}$ is a vector of length $M$, and $\mathcal{A}$ indicates a given compressive operator.

In this work, the randomly permuted Walsh-Hadamard operator~\cite{Jiang2012a} and the randomly permuted noiselet operator~\cite{noiselets2001} will be employed as compressive operators because of their low computational cost and easy hardware implementation. Compressive operator can be instantiated as $\mathcal{A}=\mathbf{D} \cdot \mathbf{H} \cdot \mathbf{P}$, where $\mathbf{P}$ is a random permutation matrix, $\mathbf{H}$ is the Walsh-Hadamard transform or the noiselet transform, and $\mathbf{D}$ is a randomly down sampling operator. As stated in \cite{CI2006b}, compressive operator often encodes video volume $\mathcal{X}_0$ through two ways.
One is the holistic manner, i.e., $\textbf{y}$= $\mathbf{D} \cdot \mathbf{H} \cdot \mathbf{P} (\mathbf{x}_0)$, which directly collects full 3D measurements of a video sequence. The other is the frame-wise manner, i.e., $\textbf{y}_d$= $\mathbf{D}_d \cdot \mathbf{H}_d \cdot \mathbf{P}_d (\mathbf{x}_{0}^{d})~(d=1,2,\cdots,D)$, which collects 2D \emph{frame-by-frame} measurements $\mathbf{y}_d$ and then concatenates all $\mathbf{y}_d$
into a long vector $\mathbf{y}$. In most experiments of this work, compressive operator $\mathcal{A}$ will be set as the frame-by-frame one.
\subsection{Reconstruction and Separation of Video Volume}
As we know, recovering $\mathcal{X}_0$ and simultaneously separating $\mathcal{X}_1$ with $\mathcal{X}_2$ from the compressive measurements $\mathbf{y}$ is a heavily ill-posed inverse problem. Hence, it is necessary to regularize this inverse problem by discovering the underlying video prior knowledge. Mathematically, the regularized inverse problem can be generally formulated as
\begin{equation}
\label{general_model}
\begin{split}
&\min_{\mathbf{x}_0,\mathbf{x}_1,\mathbf{x}_2}   \lambda \Omega_2(\mathbf{x}_2) + \Omega_1(\mathbf{x}_1) \\
&s.t.~~\mathbf{x}_0 = \mathbf{x}_2 +  \mathbf{x}_1, ~~\mathbf{y} = \mathcal{A}(\mathbf{x}_0),
\end{split}
\end{equation}
where $\Omega_1(\mathbf{x}_1)$ and $\Omega_2(\mathbf{x}_2)$ are the prior knowledge modeling terms on video background and foreground, respectively; $\mathbf{x}_0$, $\mathbf{x}_1$ and $\mathbf{x}_2$ are the vectorizations of~$\mathcal{X}_0$,~$\mathcal{X}_1$ and~$\mathcal{X}_2$, respectively; and $\lambda$ is a trade-off parameter between the terms $\Omega_1(\mathbf{x}_1)$ and $\Omega_2(\mathbf{x}_2)$.

In the following section, we will fully discover the priors for surveillance videos and characterize these priors using tensor algebra, which naturally instantiates the general model in Eq.~\eqref{general_model} into the practical models.

\section{Tensor RPCA models for the BSCM}
In this section, we will present our proposed tensor robust principal component (PCA) models for the BSCM task. We first review the basics in multi-linear algebra. Then, we present our basic model for video decomposition, and further propose detailed foreground model and background model by considering the spatio-temporal continuity of video foreground and spatial-temporal correlations of video background. In the background modeling, we propose two models that represent the video background as a single tensor and several patch-level tensors over groups of similar 3D patches respectively. We utilize tensor Tucker decomposition to model the video background in the aforementioned holistic and patch-based representations, and produce two tensor RPCA models (named H-TenRPCA and PG-TenRPCA), respectively.
\begin{table}[htp]
\renewcommand\arraystretch{1.3}
\centering
\caption{Notations}
\begin{tabular}{lp{0.05cm}p{5cm}} 
\toprule[1pt]
Notations & & Explanations \\
\hline
 $\mathcal{X}$, $\mathbf{X}$, $\mathbf{x}$, $x$   &    & tensor, matrix, vector, scalar.   \\
 $\mathbf{x}(:,i_2,i_3,\cdots,i_N)$               &    & fiber of tensor $\mathcal{X}$ obtained by fixing all but one index. \\

$\mathbf{X}(:,:,i_3,\cdots,i_N)$   &     & slice of tensor $\mathcal{X}$ obtained by fixing all but two indices. \\
$\mathbf{X}_{(n)}$ or $\mathcal{X}_{(n)}$ & &mode-$n$ matricization of tensor $\mathcal{X}$ $\in$ $\Re^{I_1\times I_2\times,\cdots,\times I_N}$, obtained by arranging the mode-$n$ fibers as the columns of the resulting matrix of size $\Re^{I_n \times \prod_{k \neq n} I_k}$. \\
$\textbf{Vec}(\mathcal{X})$ & &vectorization of tensor $\mathcal{X}.$ \\
$\textbf{Ten}(\mathbf{x})$& &tensorization of vector $\mathbf{x}$, i.e., the inverse operation of $\textbf{Vec}$.\\
$(r_1,r_2,\cdots,r_N)$  & & multi-linear rank, where $r_n = \text{Rank}(\textbf{X}_{(n)})$, $n = 1,2,\cdots,N.$ \\
$\langle \mathcal{X}, \mathcal{Y} \rangle$     &     & inner product of tensor $\mathcal{X}$ and $\mathcal{Y}$. \\
$\| \mathcal{X} \|_{F}$                        &     & Frobenius norm of tensor $\mathcal{X}$. \\
$\mathcal{Y}=\mathcal{X} \times_n \textbf{U} $ &    & mode-$n$ multiplication of $\mathcal{X}$ and $\textbf{U}$ with the matrix representation $\textbf{Y}_{(n)}= \textbf{U} \textbf{X}_{(n)}$. \\
\bottomrule[1pt]
\end{tabular}
\label{fuhao}
\end{table}
\subsection{Tensor Basics}
A tensor can be seen as a multi-index numerical array. The order of a tensor is the number of its modes or dimensions.
A real-valued tensor of order $N$ is denoted by $\mathcal{X} \in \Re^{I_1\times I_2 ... \times I_N}$ and its entries by $x_{i_1,i_2,\cdots,i_N}$.
Then an $N\times 1$ vector $\mathbf{x}$ is considered as a tensor of order one, and an $N\times M$ matrix $\textbf{X}$ as a tensor of order two.
Subtensors are parts of the original tensor, created when only a fixed subset of indices is used. Vector-valued subtensor are called fibers,
defined by fixing every index but one, and matrix valued subtensor are called slices, obtained by fixing all but two indices.
Manipulation of tensors often requires their reformatting (reshaping); a particular case of reshaping tensors into matrices is termed as matrix unfolding or matricization. The multi-linear rank of a $N$-order tensor is the tuple of the ranks of the mode-n unfoldings.
The inner product of two same-sized tensors $\mathcal{X}$ and $\mathcal{Y}$ is the sum of the products of their entries. The mode-n multiplication of a tensor $\mathcal{X}$ with a matrix $\textbf{U}$ amounts to the multiplication of all mode-$n$ vector fibers with $\textbf{U}$, i.e., $ (\mathcal{X} \times_n \textbf{U})_{i_1,i_2,\cdots,j,\cdots,i_N} =\sum_{i_n} x_{i_1,i_2,\cdots,i_n,\cdots,i_N} \cdot u_{j,i_n}$. The used tensor notations are summarized in Table~\ref{fuhao}. For more details about multi-linear algebra, please see~\cite{Kolda2009,Cichocki2009}.
\subsection{General Decomposition Model of Video Volume}
For surveillance videos in reality, we observe that there might exist some disturbances (e.g., randomly dynamic components) in the video background, for example, the fountain in the ``Fountain" video and the ripple in the ``WaterSurface'' video as shown in Fig.~\ref{realData}. Therefore, the assumption that the video background is strictly low rank may be not accurate in most existing work~\cite{Waters2011, Jiang2012a, candes2011}.

\begin{figure}[htp]
\centering
\vspace{-2mm}
\includegraphics[width=3.5in,height=1in]{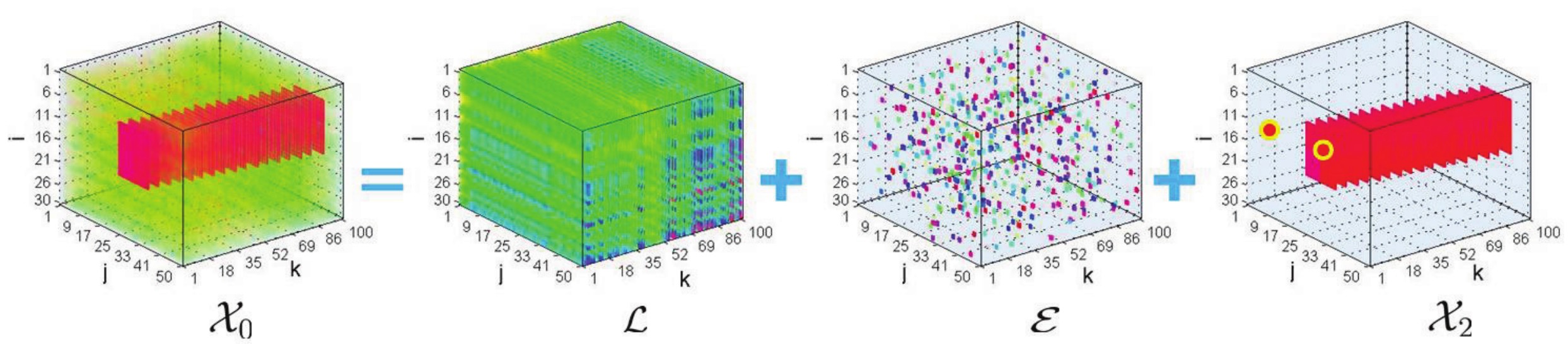}
\vspace{-6mm}
\caption{Illustration for the decomposition of video volume.}
\vspace{-5mm}
\label{decompModel}
\end{figure}

Motivated by the above observation, we further decompose the video background \big(i.e., $\mathcal{X}_1$ in Eq.~(\ref{general_model})\big) as the sum of the low rank component $\mathcal{L}$ (the ideal video background) and the disturbance $\mathcal{E}$ in this work. Then, the video volume $\mathcal{X}_0$ can be decomposed as $\mathcal{X}_0 = \mathcal{X}_2+ \mathcal{E}+ \mathcal{L}$; see Fig~\ref{decompModel}. Accordingly, the general model in Eq.~\eqref{general_model} is replaced by the following more accurate version:
\begin{equation}
\label{general_model2}
\begin{split}
&\min_{\mathbf{x}_0,\mathbf{e},\mathcal{L},\mathbf{x}_2} \lambda \Omega_2(\mathbf{x}_2) + \zeta  \Upsilon(\mathbf{e}) +  \Phi(\mathcal{L})   \\
&s.t.~~\mathbf{x}_0 = \mathbf{x}_2 + \mathbf{e} +\textbf{Vec}(\mathcal{L}), ~~\mathbf{y} = \mathcal{A}(\mathbf{x}_0),
\end{split}
\end{equation}
where $\mathbf{e}=\textbf{Vec}(\mathcal{E})$, $\lambda$ and $\zeta$ are both trade-off parameters, and $\Upsilon(\mathbf{e})$ is often specified as $\frac{1}{2}\| \mathbf{e}\|^{2}$ or $\|\mathbf{e}\|_1$. In this work, we use $\Upsilon(\mathbf{e}) = \frac{1}{2}\| \mathbf{e}\|^{2}$. In the following, we will focus on discovering the priors of video
foreground/background and then encoding these priors, i.e., specifying $\Omega_2(\mathbf{x}_2)$ and $\Phi(\mathcal{L})$.
\subsection{Foreground Modeling}
The video foreground is considered as the salient moving objects in a video, which often occupies a certain proportion of contiguous region of the video frames. For example, the car in the ``simulated" video as shown in Fig.~\ref{mdlPri}(b), and the pedestrians in the ``Hall" video as shown in Fig.~\ref{realData}. These moving objects to be detected commonly occupy a certain \textit{contiguous} region in the spatial domain. Fig.~\ref{decompModel} gives an intuitive illustration, indicating that we hope to detect $\mathcal{X}_2$ with contiguous supports in spatial domain instead of disturbance $\mathcal{E}$ with disconnected supports. Additionally, the moving trace of foreground object is temporally smooth, which can be observed from the example of car in the ``simulated" video shown in Fig.~\ref{mdlPri}(b) and the pedestrians in the ``Hall" video  shown in Fig.~\ref{realData}. We term these two discovered structures of video foreground as the~\textbf{\textit{spatio-temporal continuity prior}}.
\begin{figure}[htp]
\centering
\includegraphics[width=2.6in,height=2.2in]{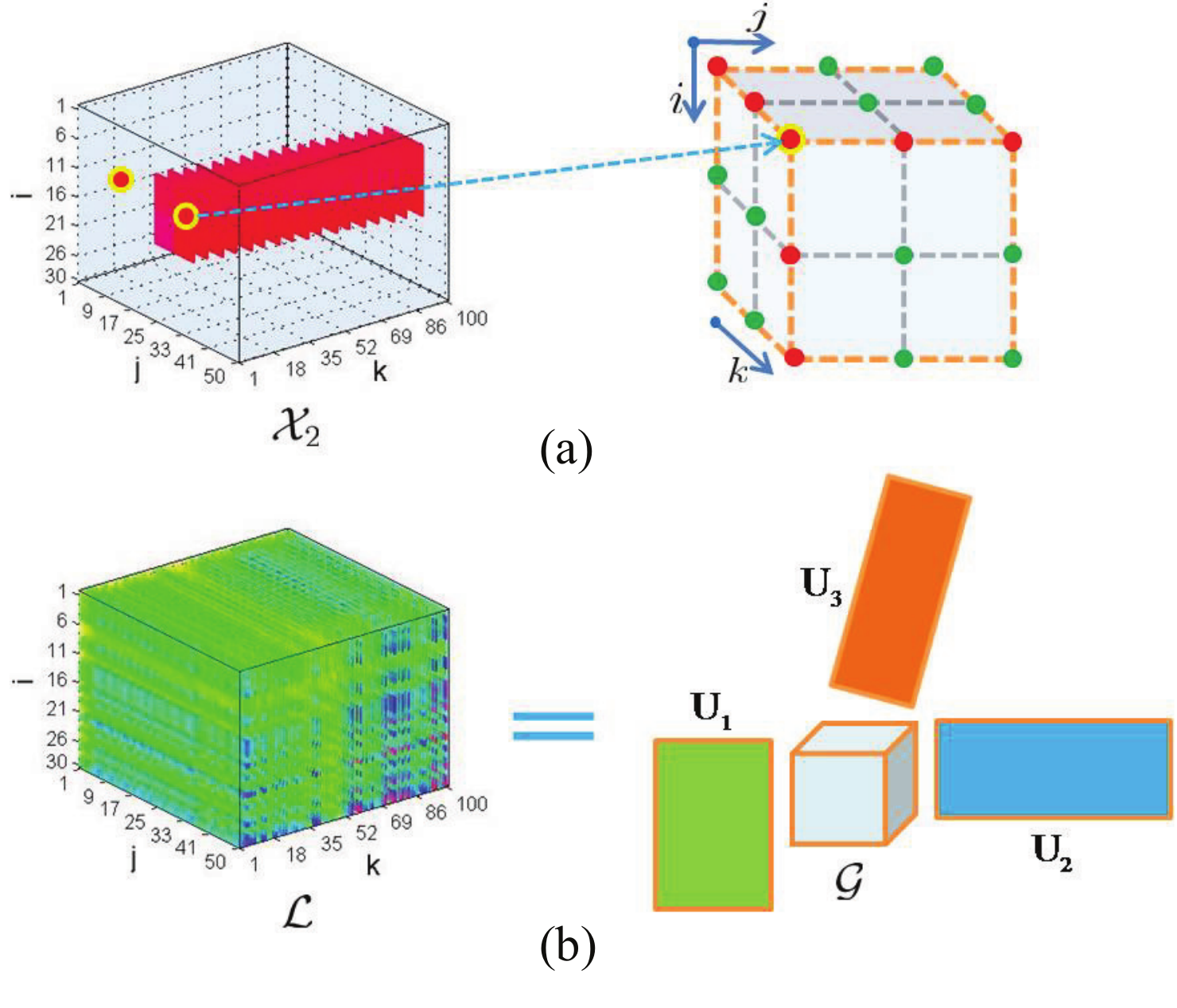}
\vspace{-3mm}
\caption{Illustration for the holistic reconstruction and separation model.
(a) 3D-TV on the voxel;
(b) The ideal video background can be reconstructed by Tucker decomposition.}
\label{fig_model1}
\end{figure}

We define a 3D total variation (TV) to model the spatio-temporal continuity. As shown in Fig.~\ref{fig_model1}(a),
for the reference voxel $(i,j,k)$ in video foreground $\mathcal{X}_2$, we devise the following quantity to describe its spatio-temporal continuity:
\begin{equation*}
\begin{split}
&\text{TV}_{i,j,k}(\mathbf{x}_2):=|\mathcal{X}_2(i,j,k) - \mathcal{X}_2(i+1,j,k)| +\\
& |\mathcal{X}_2(i,j,k) - \mathcal{X}_2(i,j+1,k)| +|\mathcal{X}_2(i,j,k) - \mathcal{X}_2(i,j,k+1)|.
\end{split}
\end{equation*}
Summing the quantity with respect to all the voxels leads to the proposed 3D-TV:
\begin{equation*}
\|\mathbf{x}_2\|_{\text{3D-TV}}: = \sum_{i,j,k}\text{TV}_{i,j,k}(\mathbf{x}_2).
\end{equation*}
It is worth noting that in this work we assume that video boundaries are processed to be circular, hence 3D-TV of the voxels in video boundaries can be defined.

For better illustration, we further introduce difference operator to rewrite $\|\mathbf{x}_2\|_{\text{3D-TV}}$. Let $\mathcal{X}(i,j,k)$ denote the intensity at the voxel $(i,j,k)$, and
\begin{equation*}
\begin{split}
& \mathcal{X}_{h}(i,j,k) : = \mathcal{X}(i,j+1,k) - \mathcal{X}(i,j,k), \\
& \mathcal{X}_{v}(i,j,k) : = \mathcal{X}(i+1,j,k) - \mathcal{X}(i,j,k), \\
& \mathcal{X}_{t}(i,j,k) : = \mathcal{X}(i,j,k+1) - \mathcal{X}(i,j,k),
\end{split}
\end{equation*}
denote three difference operations at the voxel $(i,j,k)$ along the horizontal, vertical, and temporal directions respectively.
We can now easily introduce three difference operators with respect to three different direction as follows:
\begin{equation*}
\mathbf{D}_h\mathbf{x} :=\textbf{Vec}(\mathcal{X}_h),~~\mathbf{D}_v\mathbf{x} :=\textbf{Vec}(\mathcal{X}_v),
~~\mathbf{D}_t\mathbf{x} :=\textbf{Vec}(\mathcal{X}_t),
\end{equation*}
where $\mathbf{x}=\textbf{Vec}(\mathcal{X})$.
Let~$\mathbf{D}\mathbf{x} : = [(\mathbf{D}_h\mathbf{x})^T, (\mathbf{D}_v\mathbf{x})^T, (\mathbf{D}_t\mathbf{x})^T]^{T}$ denote the concatenation of three difference operations. It is easy to see that 3D-TV amounts to $\ell_1$ norm of the difference vectors:
\begin{equation}
\label{fgpri}
\begin{split}
\|\mathbf{x}_2 \|_{\text{3D-TV}} &= \| \mathbf{D}\mathbf{x}_2\|_1  \\
&= \| \mathbf{D}_h\mathbf{x}_2\|_1 + \| \mathbf{D}_v\mathbf{x}_2\|_1 + \| \mathbf{D}_t\mathbf{x}_2\|_1.
\end{split}
\end{equation}
\subsection{Background Modeling}
\subsubsection{Holistic Background Modeling}
As discussed in the introduction part, video background within a short period possesses the spatio-temporal correlation.
The strong temporal correlation in video background implies that matrix unfolding $\mathbf{X_1}_{(3)}$ in the temporal mode can be approximated by a low rank matrix. Mathematically, $\mathbf{X_1}_{(3)} = \mathbf{U}_3 \mathbf{C}_3 + \mathbf{E}_{(3)}$, where $\mathbf{U}_3$ is a low rank matrix of rank $r_3\ll D$ and $\mathbf{E}_{(3)}$ is the disturbance. The weak spatial correlation in video background implies that the matrix unfoldings $\mathbf{X_1}_{(1)}$ and $\mathbf{X_1}_{(2)}$ in the height and width modes can be approximated by two high rank matrices, respectively.
Mathematically, $\mathbf{X_1}_{(1)} = \mathbf{U}_1 \mathbf{C}_1 + \mathbf{E}_{(1)}$ and $\mathbf{X_1}_{(2)} =  \mathbf{U}_2 \mathbf{C}_2 + \mathbf{E}_{(2)}$, where $\mathbf{U}_1$ and $\mathbf{U}_2$ are both two high rank matrices of rank $r_1<H$ and $r_2<W$, respectively.
Resorting to the well-known Tucker decomposition in multi-linear algebra, the matrix factorizations above can be aggregated together as follows:
\begin{equation}
\label{bgpri1}
\mathcal{X}_1 = \mathcal{G} \times_1 \mathbf{U}_1 \times_2 \mathbf{U}_2 \times_3 \mathbf{U}_3 + \mathcal{E},
\end{equation}
where factor matrices $\textbf{U}_1$ and $\textbf{U}_2$ are orthogonal in columns for two spatial modes, factor matrix $\textbf{U}_3$ is orthogonal in columns for temporal mode, core tensor $\mathcal{G}$ interacts these factors, and $\mathcal{E}$ is the disturbance. Let $\mathcal{L} = \mathcal{G} \times_1 \mathbf{U}_1 \times_2 \mathbf{U}_2 \times_3 \mathbf{U}_3$. We call $\mathcal{L}$ the ideal video background. Our holistic background modeling is intuitively illustrated in Fig.~\ref{fig_model1}(b).

Compared to matrix modeling technique, the advantage of tensor modeling technique is that it can not only characterize the temporal correlation but also the spatial correlation in video background. Thus it can reconstruct more accurate video background.
\subsubsection{Patch-based Background Modeling}
Patch-based modeling is a popular and local style modeling technique and widely used in the community of image processing. \textit{Nonlocal self-similarity}~\cite{Mairal2009,Dong_b,Dong_d,Jian2011,Peng2014} is a patch-based powerful prior and means that one patch
in one image has many similar\footnote{Here, two patches are defined as similar if the Euclidean distance between two patch vectors is smaller than a given threshold.} structure patches. The similarity of patches implies the correlation of patches. In this work, we will extend this prior into 3D case and approximately reconstruct video background $\mathcal{X}_1$ (or say, accurately reconstruct the ideal video background $\mathcal{L}$) through modeling the video background by groups of similar video 3D patches, where each patch group corresponds to a tensor.

\begin{figure}[htp]
\centering\vspace{-2mm}
\includegraphics[width=3.2in,height=1.8in]{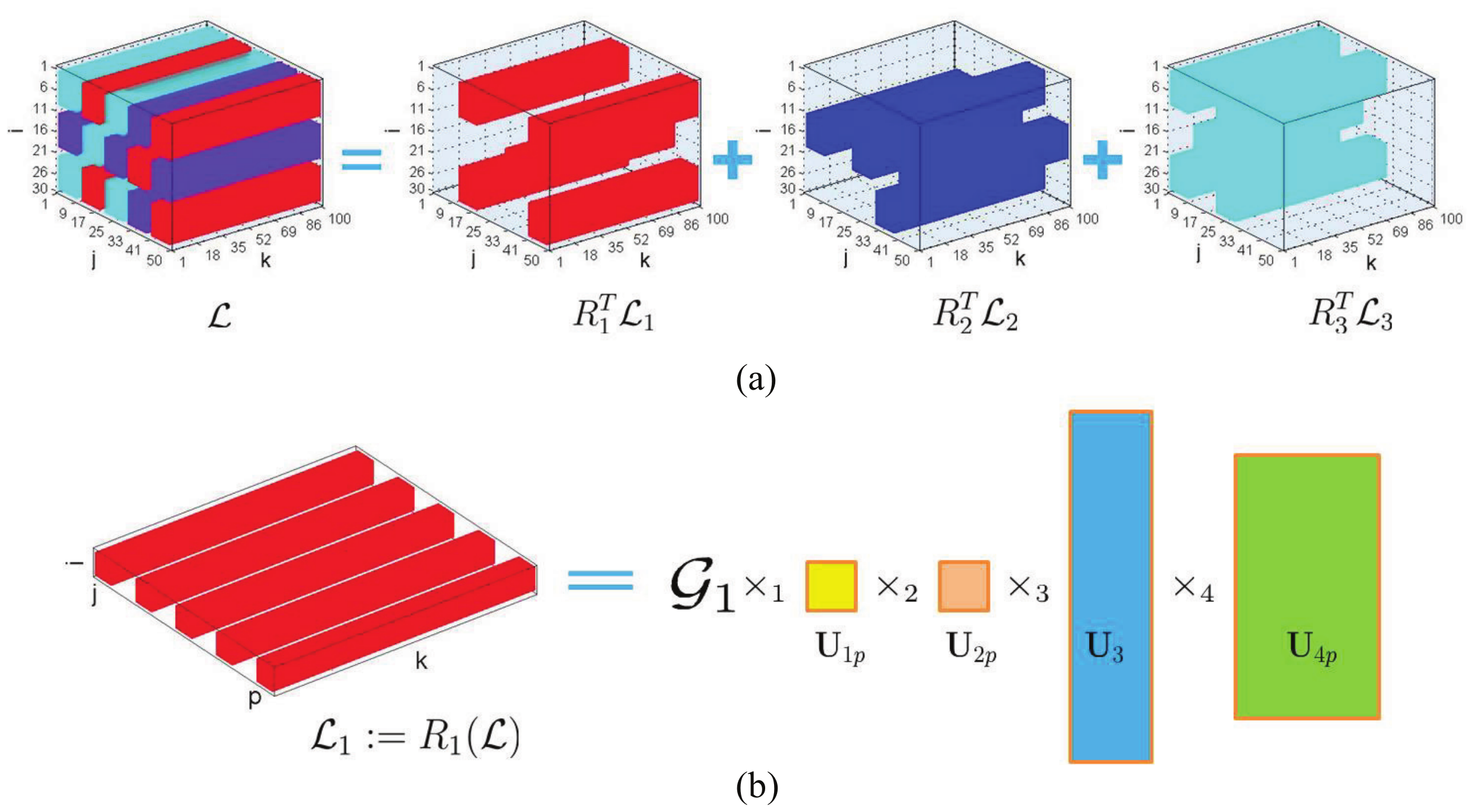}
\vspace{-2mm}
\caption{Illustration for the patch-based background modeling.
(a) The non-overlapped patches on the ideal video background can be clustered into three clusters;
(b) The 4-order tensor composed of each cluster can be reconstructed by low rank Tucker decomposition.}
\label{fig_model2}
\end{figure}

Specifically, we firstly segment video background $\mathcal{X}_1$ into many overlapped 3D patches of the size $w \times w \times D$ and then collect these 3D patches as a patch set $\mathcal{S}$:
$
\mathcal{S}= \{ \mathcal{P}_i \in \Re^{w\times w \times D}: i \in \Gamma \},$ where $\Gamma$ indicates the index set and $\mathcal{P}_i$ is the $i$-th 3D patch in the set.
These 3D patches are commonly similar to each other; see Fig.~\ref{mdlPri}(a) for an example. We cluster\footnote{The technical details concerning how to cluster will be stated in the subsequent subsection,~\textit{Implementation Issues}.} the patch set $\mathcal{S}$ into $K$ clusters and then collect each cluster as a 4-order tensor.
Mathematically, let $\mathbf{C}_{p}^{o}$ be a matrix extracting the $o$-th 3D patch in the $p$-th cluster as a vector of the size $(w^2D)\times 1$,
and define $\mathbf{R}_p\mathbf{x}_1$ as:
\begin{equation}
\mathbf{R}_p\mathbf{x}_1:= \left(
 \begin{array}{c}
 \mathbf{C}_{p}^{1}\mathbf{x}_1 \\
 \mathbf{C}_{p}^{2}\mathbf{x}_1 \\
  \vdots \\
  \mathbf{C}_{p}^{N}\mathbf{x}_1 \\
 \end{array} \right),
\end{equation}
where $N$ is the number of 3D patches in the $p$-th cluster. Then $\mathbf{R}_p\mathbf{x}_1$ can be reshaped into a 4-order tensor $\textbf{Ten}(\mathbf{R}_p\mathbf{x}_1)$ of the size $w\times w\times D \times N$, denoted by $\mathcal{R}_p(\mathcal{X}_1)$\footnote{$\mathcal{R}_{p}$ indicates the operation which first extracts all 3D patches in the $p$-cluster from the video volume, and then arranges these 3D patches as a 4-order tensor, i.e., $\mathcal{R}_p(\mathcal{X}_1) = \textbf{Ten}\big(\mathbf{R}_p\textbf{Vec}(\mathcal{X}_1)\big)$ while $\mathcal{R}_{p}^{T}$ indicates its inverse-order operation,  i.e., $\mathcal{R}_{p}^{T}(\mathcal{L}_p) = \textbf{Ten}\big(\mathbf{R}_{p}^{T}\textbf{Vec}(\mathcal{L}_p)\big)$; see Fig.~\ref{fig_model2}.}; see Fig.~\ref{fig_model2} for an intuitive illustration.
Because the patches in each cluster have very similar structures,
$\textbf{Ten}(\mathbf{R}_p\mathbf{x}_1)$ can then be expectedly approximated by a low rank tensor $\mathcal{L}_p$, i.e., $\textbf{Ten}(\mathbf{R}_p\mathbf{x}_1)\approx \mathcal{L}_p$.
The modeling of $\mathcal{L}_p$ will be determined shortly. Then, the clean and ideal video background can be estimated by
solving the following optimization problem: $$\min_{\mathbf{x}_1} \sum_{p=1}^{K} \| \mathbf{R}_p(\mathbf{x}_1) - \textbf{Vec}(\mathcal{L}_p)\|^2.$$
The solution of this optimization problem can be easily derived as $\widehat{\mathbf{x}}_1= (\sum_p \mathbf{R}_{p}^{T} \mathbf{R}_{p})^{-1} \sum_p \mathbf{R}_{p}^{T} \textbf{Vec}(\mathcal{L}_p)$. Let us denote $\textbf{Ten}(\widehat{\mathbf{x}}_1)$ by $\mathcal{L}$. Hence, $\mathcal{L}$ can be represented as:
\begin{equation}
\label{aggre}
\begin{split}
\mathcal{L} &= \textbf{Ten}\big((\sum_p \mathbf{R}_{p}^{T} \mathbf{R}_{p})^{-1}\sum_p \mathbf{R}_{p}^{T} \textbf{Vec}(\mathcal{L}_p)\big),\\
\end{split}
\end{equation}
which means that the ideal video background $\mathcal{L}$ can be obtained by summing all clusters followed by an averaging operation. When the patches in the patch set $\mathcal{S}$ are not overlapped, $(\sum_p \mathbf{R}_{p}^{T} \mathbf{R}_{p})^{-1}$ reduces to an identify matrix. Fig.~\ref{fig_model2}(a) illustrates this procedure in which the averaging operation is not required.

$\mathcal{L}_p$ is one 4-order tensor of the size $w\times w \times D \times N$ which collects all 3D patches in the $p$-th cluster. Because the ideal video background possesses a strong correlation among the frames, $(\mathcal{L}_p)_{(3)}$ is low rank. Moreover, the observation that the patches in each cluster have very similar structures implies that $(\mathcal{L}_p)_{(4)}$ is also low rank. Combining these two points, we can likewise model $\mathcal{L}_p$ by Tucker decomposition:
\begin{displaymath}
\mathcal{L}_p = \mathcal{G}_p \times_1 \mathbf{U}_{1p} \times_2 \mathbf{U}_{2p} \times \mathbf{U}_3 \times \mathbf{U}_{4p},
\end{displaymath}
where $\mathcal{G}_p$ is core tensor, and $\mathbf{U}_{1p}$, $\mathbf{U}_{2p}$, $\mathbf{U}_{3}$ and $\mathbf{U}_{4p}$ are factor matrices orthogonal in columns.
Note that the factor matrices in the temporal mode for all $p$ are set as a shared matrix $\mathbf{U}_{3}$, insuring that $\mathcal{L}$ is low rank
in the temporal mode on the whole. Fig.~\ref{fig_model2}(b) gives an intuitive illustration. The video background now can be modeled as:
\begin{equation}
\label{bgpri2}
\begin{split}
\mathcal{X}_1 &= \mathcal{L}+ \mathcal{E} \\
              &= \textbf{Ten}\big((\sum_p \mathbf{R}_{p}^{T} \mathbf{R}_{p})^{-1}\sum_p \mathbf{R}_{p}^{T} \textbf{Vec}(\mathcal{L}_p)\big) + \mathcal{E},
\end{split}
\end{equation}
where factor matrices $\mathbf{U}_{jp}~(j=1,2,4)$ and $\mathbf{U}_3$ are orthogonal in columns.
\subsection{Reconstruction and Separation Models}
We now can instantiate the general model in Eq.~\eqref{general_model2}. Integrating the modelings of video foreground in Eq.~\eqref{fgpri} and video background in Eq.~\eqref{bgpri1} into the general model in Eq.~\eqref{general_model2} leads to the following holistic TenRPCA model (\textbf{\emph{H-TenRPCA}}):
\begin{equation}
\label{rs_model1}
\begin{split}
&\min_{ \begin{subarray}{c} \mathbf{x}_0, \mathbf{x}_2,\mathbf{e}, \\
                    \mathcal{G}, \mathbf{U}_{j}\\
         \end{subarray}
         } \lambda \|\mathbf{D} \mathbf{x}_2 \|_1 + \frac{1}{2}\|\mathbf{e}\|^{2}  \\
&s.t.~\mathbf{x}_0 = \mathbf{x}_2 + \mathbf{e} +  \textbf{Vec}(\mathcal{G} \times_1 \mathbf{U}_1 \times_2 \mathbf{U}_2 \times_3 \mathbf{U}_3),  \\
&~~~~~ \mathbf{y} = \mathcal{A}(\mathbf{x}_0),\\
\end{split}
\end{equation}
where the factor matrices $\mathbf{U}_{j}~(j=1,2,3)$ are orthogonal in columns.

Likewise, integrating the modeling of video foreground in Eq.~\eqref{fgpri}  and the patch-based modeling of video background in Eq.~\eqref{bgpri2} leads to the following patch-group-based tensor RPCA model (\textbf{\emph{PG-TenRPCA}}):
\begin{equation}
\label{rs_model2}
\begin{split}
&\min_{ \begin{subarray}{c} \mathbf{x}_0, \mathbf{x}_2,\mathbf{e}, \\
                    \mathcal{G}_{p}, \mathbf{U}_{jp}, \mathbf{U}_3 \\
         \end{subarray}
         }\lambda \|\mathbf{D} \mathbf{x}_2 \|_1+ \frac{1}{2}\|\mathbf{e}\|^{2} \\
&s.t.~\mathbf{x}_0 =\mathbf{x}_2 + \mathbf{e}~+ (\sum_p \mathbf{R}_{p}^{T} \mathbf{R}_{p})^{-1} \sum_p \mathbf{R}_{p}^{T} \textbf{Vec}\\
&~~~~~~~~~~~~(\mathcal{G}_p \times_1 \mathbf{U}_{1p} \times_2 \mathbf{U}_{2p} \times_3 \mathbf{U}_3 \times_4 \mathbf{U}_{4p}), \\
&~~~~~ \mathbf{y} = \mathcal{A}(\mathbf{x}_0),
\end{split}
\end{equation}
where the factor matrices $\mathbf{U}_{jp}~(j=1,2,4)$ and $\mathbf{U}_3$ are orthogonal in columns.

In the following section, we will design efficient algorithms to solve the proposed models. Note that these models are non-convex, and therefore, we can only wish to find local solutions.
\section{Optimization Algorithms}
In this section, we first develop an efficient algorithm based on ADMM for solving the proposed model of H-TenRPCA in Eq.~\eqref{rs_model1}. Then, the algorithm is slightly modified to solve the PG-TenRPCA model in Eq.~\eqref{rs_model2}. Finally, we present the implementation details of our optimization algorithms.
\subsection{Optimization Algorithm for H-TenRPCA}
We optimize the H-TenRPCA model using a multi-block version of the alternating direction method of multipliers (ADMM)~\cite{Boyd2012,Deng2012,Mota2013,Shen2014,ZQLuo2013,Daniel2015}. The H-TenRPCA model in Eq.~\eqref{rs_model1} can be rewritten as the following equivalent form:
\begin{equation}
\label{rs_model11}
\begin{split}
&\min_{ \begin{subarray}{c} \mathbf{x}_0, \mathbf{x}_2,\mathbf{e}, \\
                    \mathbf{f},\mathcal{G}, \mathbf{U}_{j}\\
         \end{subarray}
         } \lambda \|\mathbf{f} \|_1 + \frac{1}{2}\|\mathbf{e}\|^{2}  \\
&s.t.~ \mathbf{f}  = \mathbf{D} \mathbf{x}_2 ,\\
&~~~~~ \mathbf{x}_0 = \mathbf{x}_2 + \mathbf{e} +  \textbf{Vec}(\mathcal{G} \times_1 \mathbf{U}_1 \times_2 \mathbf{U}_2 \times_3 \mathbf{U}_3),  \\
&~~~~~ \mathbf{y} = \mathcal{A}(\mathbf{x}_0),
\end{split}
\end{equation}
where the factor matrices $\mathbf{U}_{j}$~($j = 1,2,3$) are orthogonal in columns. This constrained optimization problem can be solved by its Lagrangian dual form. The augmented Lagrangian function of problem in Eq.~\eqref{rs_model11} can be written as:
\begin{equation*}
\begin{split}
&L_A(\mathbf{x}_0, \mathcal{G}, \mathbf{U}_i,\mathbf{e},\mathbf{x}_2,\mathbf{f}) =\lambda \|\mathbf{f} \|_1 + \frac{1}{2}\|\mathbf{e}\|^{2}\\
&- \langle \bm{\lambda}^{\mathbf{f}}, \mathbf{f} - \mathbf{D}\mathbf{x}_{2} \rangle + \frac{\beta^{\mathbf{f}}}{2} \| \mathbf{f} - \mathbf{D}\mathbf{x}_{2} \|^2 \\
&-\langle \bm{\lambda}^{\mathbf{x}_0}, \mathbf{x}_0 - \mathbf{x}_2 - \mathbf{e} -  \textbf{Vec}(\mathcal{G} \times_1 \mathbf{U}_1 \times_2 \mathbf{U}_2 \times_3 \mathbf{U}_3) \rangle \\
&+ \frac{\beta^{\mathbf{x}_0}}{2} \| \mathbf{x}_0 - \mathbf{x}_2 - \mathbf{e} - \textbf{Vec}( \mathcal{G} \times_1 \mathbf{U}_1 \times_2 \mathbf{U}_2 \times_3 \mathbf{U}_3) \|^{2} \\
& -\langle \bm{\lambda}^{\mathbf{y}},\mathbf{y}-\mathcal{A}(\mathbf{x}_0)\rangle + \frac{\beta^\mathbf{y}}{2} \| \mathbf{y} - \mathcal{A}(\mathbf{x}_0) \|^2,
\end{split}
\end{equation*}
where $\bm{\lambda}^{\mathbf{f}}$, $\bm{\lambda}^{\mathbf{x}_0}$ and $\bm{\lambda}^{\mathbf{y}}$ are the Lagrange multiplier vectors,
and $\beta^{\mathbf{f}}$, $\beta^{\mathbf{x}_0}$ and $\beta^{\mathbf{y}}$ are positive penalty scalars.
It is difficult to simultaneously optimize all these variables.  We therefore approximately solve this optimization problem by alternatively minimizing one variable with the others fixed. This procedure is the so-called multi-block alternating direction method of multiples (ADMM). Under the framework of multi-block ADMM, the optimization problem of $L_A$ with respect to each variable can be solved by the following sub-problems:
\subsubsection{$\mathbf{x}_0$ sub-problem}
Optimizing $L_A$ with respect to $\mathbf{x}_0$ can be treated as
solving the following linear system:
\begin{equation*}
\begin{split}
&(\beta^{\mathbf{x}_0} \mathbf{I} + \beta^{\mathbf{y}} \mathcal{A}^{*} \mathcal{A} ) \mathbf{x}_0 =\\
& \bm{\lambda}^{\mathbf{x}_0} + \beta^{\mathbf{x}_0}\big(\mathbf{x}_2 + \mathbf{e} + \textbf{Vec}(\mathcal{L})\big) + \mathcal{A}^{*}(\beta^{\mathbf{y}} \mathbf{y} - \bm{\lambda}^{\mathbf{y}}),
\end{split}
\end{equation*}
where $\mathcal{A}^{*}$ indicates the adjoint of $\mathcal{A}$ and $\mathcal{L} = \mathcal{G} \times_1 \mathbf{U}_1 \times_2 \mathbf{U}_2 \times_3 \mathbf{U}_3$. Obviously, this linear system can be solved by off-the-shelf conjugate gradient techniques. When $\mathcal{A}\mathcal{A}^{*} = \mathbf{I}$, this linear system has the following closed-form solution:
\begin{equation}
\label{M1_sub1}
\mathbf{x}_0 = ( \mathbf{I} - \frac{ \beta^{\mathbf{y}} }{ \beta^{\mathbf{x}_0} +\beta^{\mathbf{y}} } \mathcal{A}^{*} \mathcal{A}  ) \frac{\mathbf{c}^{\mathbf{y}}}{\beta^{\mathbf{x}_0}},
\end{equation}
where $\mathbf{c}^{\mathbf{y}}=\bm{\lambda}^{\mathbf{x}_0} + \beta^{\mathbf{x}_0}\big(\mathbf{x}_2 + \mathbf{e} + \textbf{Vec}(\mathcal{L})\big) + \mathcal{A}^{*}(\beta^{\mathbf{y}}\mathbf{y} - \bm{\lambda}^{\mathbf{y}})$.
\subsubsection{$\mathcal{G}$ and $\mathbf{U}_i$ sub-problems} The optimization sub-problem of $L_A$ with respect to $\mathcal{G}$ and $\mathbf{U}_i~(i=1,2,3)$ can be rewritten as:
\begin{equation}
\label{M1_sub2}
\min_{\mathcal{G}, \mathbf{U}_i} \frac{1}{2} \| \widetilde{\mathcal{X}_1} -\mathcal{G} \times_1 \mathbf{U}_1 \times_2 \mathbf{U}_2 \times \mathbf{U}_3 \|_{F}^2~~~ s.t.~~ \mathbf{U}_i^T \mathbf{U}_i = \mathbf{I},
\end{equation}
where $\widetilde{\mathcal{X}_1} = \mathcal{X}_0  - \mathcal{X}_2 - \mathcal{E} -\textbf{Ten}(\frac{\bm{\lambda}^{\mathbf{x}_0}}{\beta^{\mathbf{x}_0}})$.
This sub-problem can be solved by the classic HOOI algorithm~\cite{Kolda2009,Cichocki2009}.
\subsubsection{$\mathbf{e}$ sub-problem}
The sub-problem of $L_A$ with respect to $\mathbf{e}$ can be solved by
\begin{equation}
\label{M1_sub3}
\mathbf{e} = \frac{ \beta^{\mathbf{x}_0 } \big(\mathbf{x}_0 - \mathbf{x}_2 - \textbf{Vec}(\mathcal{L}) - \frac{\bm{\lambda}^{\mathbf{x}_0}}{\beta^{\mathbf{x}_0}}\big)}{1 + \beta^{\mathbf{x}_0} },
\end{equation}
where $\mathcal{L} = \mathcal{G} \times_1 \mathbf{U}_1 \times_2 \mathbf{U}_2 \times_3 \mathbf{U}_3$.
\subsubsection{$\mathbf{x}_2$ sub-problem}
The sub-problem of $L_A$ with respect to $\mathbf{x}_2$ can be solved by the following linear system:
\begin{displaymath}
(\beta^{\mathbf{x}_0} \mathbf{I} + \beta^{\mathbf{f}} \mathbf{D}^{*} \mathbf{D}) \mathbf{x}_2 = \beta^{\mathbf{x}_0}\big(\mathbf{x}_0 - \textbf{Vec}(\mathcal{L}) - \mathbf{e}\big) - \bm{\lambda}^{\mathbf{x}_0} + \mathbf{D}^{*}(\beta^{\mathbf{f}}\mathbf{f} - \bm{\lambda}^{\mathbf{f}}),
\end{displaymath}
where $\mathbf{D}^*$ indicates the adjoint of $\mathbf{D}$. Let $\mathcal{C}_{\mathbf{b}} = \textbf{Ten}\big(\beta^{\mathbf{x}_0}(\mathbf{x}_0 - \textbf{Vec}(\mathcal{L}) - \mathbf{e}) - \bm{\lambda}^{\mathbf{x}_0} + \mathbf{D}^{*}(\beta^{\mathbf{f}}\mathbf{f} - \bm{\lambda}^{\mathbf{f}})\big)$.
Thanks to the block-circulant structure of the matrix corresponding to the operator $\mathbf{D}^{*} \mathbf{D}$, it can be diagonalized by the 3D FFT matrix. Therefore, $\mathcal{X}_2$ can be fast computed by
\begin{equation}
\label{M1_sub4}
\text{ifftn}\bigg( \frac{\text{fftn}({\mathcal{C}_{\mathbf{b}}})}{ \beta^{\mathbf{x}_0}\textbf{1} + \beta^{\mathbf{f}}( |\text{fftn}(\mathbf{D}_h)|^2 + |\text{fftn}(\mathbf{D}_v)|^2 + |\text{fftn}(\mathbf{D}_t)|^2) } \bigg),
\end{equation}
where $\text{fftn}$ and $\text{ifftn}$ respectively indicate fast 3D Fourier transform and its inverse transform, $|\cdot|^2$ is the element-wise square, and the division is also performed element-wisely. Note that the denominator in the equation can be pre-calculated outside the main loop, avoiding the extra computational cost.
\subsubsection{$\mathbf{f}$ sub-problem} The sub-problem of $L_A$ with respect to $\mathbf{f}$ can be rewritten as
\begin{equation*}
\min_{\mathbf{f}} \lambda \| \mathbf{f}\|_1 + \frac{\beta^{\mathbf{f}}}{2} \| \mathbf{f} - (\mathbf{D}\mathbf{x}_2 + \frac{\bm{\lambda}^{\mathbf{f}}}{\beta^{\mathbf{f}}})\|^{2},
\end{equation*}
This sub-problem can be solved by the well-known soft shrinkage operator as follows:
\begin{equation}
\label{M1_sub5}
\mathbf{f} = \text{soft}(\mathbf{D}\mathbf{x}_2 + \frac{\bm{\lambda}^{\mathbf{f}}}{\beta^{\mathbf{f}}}, \frac{\lambda}{\beta^{\mathbf{f}}}),
\end{equation}
where $\text{soft}(\mathbf{a},\tau): = \text{sgn}(\mathbf{a})\cdot\text{max}(|\mathbf{a}|-\tau, 0)$.
\subsubsection{updating multipliers}According to the ADMM, the multipliers associated with $L_A$ are updated by the following formulas:
\begin{equation}
\label{M1_sub6}
\left\{
\begin{array}{l}
\bm{\lambda}^{\mathbf{f}}\leftarrow \bm{\lambda}^{\mathbf{f}} - \gamma\beta^{\mathbf{f}} (\mathbf{f} - \mathbf{D}\mathbf{x}_2) \\
\bm{\lambda}^{\mathbf{x}_0}\leftarrow \bm{\lambda}^{\mathbf{x}_0} - \gamma\beta^{\mathbf{x}_0} \big(\mathbf{x}_0 - \textbf{Vec}(\mathcal{L}) - \mathbf{e} - \mathbf{x}_2 \big)\\
\bm{\lambda}^{\mathbf{y}}\leftarrow \bm{\lambda}^{\mathbf{y}}-\gamma\beta^\mathbf{y}\big( \mathbf{y} -\mathcal{A}(\mathbf{x}_0)\big),
\end{array}
\right.
\end{equation}
where $\gamma$ is a parameter associated with convergence rate with the value, e.g., 1.1, and the penalty parameters $\beta^{\mathbf{f}}$, $\beta^{\mathbf{x}_0}$ and $\beta^\mathbf{y}$ follow an adaptive updating scheme. Take $\beta^\mathbf{y}$ as an example.
Let $nRes=\|\mathbf{y} - \mathcal{A}( \mathbf{x}_{0}^{k} )\|$ and $nRes_{pre}$ the value of last iteration.
$\beta^{\mathbf{y}}$ is initialized by a small value $\frac{1e^{-5}}{mean(abs(\mathbf{y}))}$ and then updated by the scheme:
\begin{equation}
\label{M1_sub7}
\beta^\mathbf{y} \leftarrow c_1 \cdot \beta^\mathbf{y}~~ \text{if } nRes > c_2 \cdot nRes_{pre},
\end{equation}
where $c_1$ and $c_2$ can be taken as 1.15 and 0.95, respectively.

Let us denote the rank constraint of $\mathbf{U}_1$, $\mathbf{U}_2$ and $\mathbf{U}_3$ by $r_1$, $r_2$ and $r_3$.
The proposed algorithm for H-TenRPCA can now be summarized in Algorithm~\ref{alg1}.
\begin{algorithm}[H]
\caption{Optimization algorithm for H-TenRPCA.} \label{alg1}
\begin{algorithmic}[1]
\REQUIRE The measurements $\mathbf{y}$; The algorithm parameters: $r_3$ and $\lambda$.
\renewcommand{\algorithmicrequire}{\textbf{Initialization:}}
\REQUIRE $r_1=\text{ceil}(H\times 0.65)$ and $r_2=\text{ceil}(W\times 0.65)$;
$\mathcal{L}$ is initialized by $(r_1,r_2,r_3)$-Tucker approximation of $\textbf{Ten}(\mathcal{A}^{*}(\mathbf{y}))$;
$\mathbf{x}_2 = \mathcal{A}^{*}(\mathbf{y}) - \textbf{Vec}(\mathcal{L})$;
Other variables are initialized by $\mathbf{0}$.
\ENSURE $\mathbf{x}_0$, $\mathbf{x}_2$, and $\mathbf{x}_1=\textbf{Vec}(\mathcal{L})$.
\WHILE { not converged }
\STATE Updating $\mathbf{x}_0$ via Eq.~\eqref{M1_sub1};
\STATE Updating $\mathcal{G}$ and $\mathbf{U}_i$ or $\mathcal{L}$ via Eq.~\eqref{M1_sub2};
\STATE Updating $\mathbf{e}$ via Eq.~\eqref{M1_sub3};
\STATE Updating $\mathbf{x}_2$ via Eq.~\eqref{M1_sub4};
\STATE Updating $\mathbf{f}$ via Eq.~\eqref{M1_sub5};
\STATE Updating multipliers and the related parameters via Eqs.~\eqref{M1_sub6} and~\eqref{M1_sub7}.
\ENDWHILE
\end{algorithmic}
\end{algorithm}
\subsection{Optimization Algorithm for PG-TenRPCA}
We now slightly modify the \textbf{Algorithm}~\ref{alg1} to solve the PG-TenRPCA model in Eq.~\eqref{rs_model2}. The major modification is that the sub-problem in Eq.~\eqref{M1_sub2} is replaced by the following optimization problem:
\begin{equation*}
\label{M2_sub1}
\begin{split}
&\min_{\begin{subarray}{c}\mathcal{G}_p, \\
                     \mathbf{U}_{jp},\mathbf{U}_{3}\\
\end{subarray}}
\frac{1}{2} \| \widetilde{\mathcal{X}_1} -  \textbf{Ten}\big((\sum_p \mathbf{R}_{p}^{T} \mathbf{R}_{p})^{-1}\sum_p \mathbf{R}_{p}^{T} \textbf{Vec}(\mathcal{G}_p \times_1\mathbf{U}_{1p}\\
& ~~~~~~~~~~~~~~~~~~~~~~~~~~~~~~~~~~~~\times_2\mathbf{U}_{2p} \times_3 \mathbf{U}_3 \times_4 \mathbf{U}_{4p})\big) \|_{F}^2\\
&s.t.~~\mathbf{U}_{jp}^T \mathbf{U}_{jp} = \mathbf{I}~(j=1,2,4),~~\mathbf{U}_{3}^T \mathbf{U}_{3} = \mathbf{I}.
\end{split}
\end{equation*}
This optimization problem can be converted to the following optimization problem:
\begin{equation}
\label{M2_sub2}
\begin{split}
&\min_{ \begin{subarray}{c}\mathcal{G}_p, \\
                     \mathbf{U}_{jp},\mathbf{U}_{3}\\
\end{subarray}}
\sum_{p=1}^K \frac{1}{2} \| \mathcal{R}_p(\widetilde{\mathcal{X}_1}) -  \mathcal{G}_p \times_1 \mathbf{U}_{1p}\times_2 \mathbf{U}_{2p} \times_3 \mathbf{U}_3 \times_4 \mathbf{U}_{4p}\|_{F}^2\\
&s.t.~~\mathbf{U}_{jp}^T \mathbf{U}_{jp} = \mathbf{I}~(j=1,2,4),~~\mathbf{U}_{3}^T \mathbf{U}_{3} = \mathbf{I},
\end{split}
\end{equation}
where $\mathcal{R}_p(\widetilde{\mathcal{X}_1})=\textbf{Ten}(\mathbf{R}_p\widetilde{\mathbf{x}}_1)$ and $\widetilde{\mathbf{x}}_1$ is the vectorization of~$\widetilde{\mathcal{X}}_1$.

The optimization problem in Eq.~\eqref{M2_sub2} can be approximately solved by alternatively updating the following formulas:
\begin{align}
&\mathcal{G}_p=\mathcal{R}_p(\widetilde{\mathcal{X}}_1) \times_1 \mathbf{U}_{1p}^{T} \times_2 \mathbf{U}_{2p}^{T} \times_3 \mathbf{U}_{3}^{T} \times_4 \mathbf{U}_{4p}^{T}  \label{alg2:eq1}\\
&\mathbf{U}_{1p}=\text{SVD}\big((\mathcal{R}_p(\widetilde{\mathcal{X}_1})\times_2 \mathbf{U}_{2p}^{T} \times_3 \mathbf{U}_{3}^{T} \times_4 \mathbf{U}_{4p}^{T})_{(1)},r_1\big)  \label{alg2:eq2}\\
&\mathbf{U}_{2p}=\text{SVD}\big((\mathcal{R}_p(\widetilde{\mathcal{X}_1})\times_1 \mathbf{U}_{1p}^{T} \times_3 \mathbf{U}_{3}^{T} \times_4 \mathbf{U}_{4p}^{T})_{(2)},r_2\big)  \label{alg2:eq3}\\
&\mathbf{U}_{4p}=\text{SVD}\big((\mathcal{R}_p(\widetilde{\mathcal{X}_1})\times_1 \mathbf{U}_{1p}^{T} \times_2 \mathbf{U}_{2p}^{T}\times_3 \mathbf{U}_{3}^{T} )_{(4)},r_4\big)  \label{alg2:eq4}\\
&\mathbf{U}_{3}=\text{eigs}\big(\sum_{p=1}^{K}\mathbf{Z}_p\mathbf{Z}_{p}^{T},r_3\big),
\label{alg2:eq5}
\end{align}
where $\mathbf{Z}_p = \big(\mathcal{R}_p(\widetilde{\mathcal{X}_1}) \times_1 \mathbf{U}_{1p}^{T} \times_2 \mathbf{U}_{2p}^{T} \times_4 \mathbf{U}_{4p}^{T}\big)_{(3)}$, $\text{SVD}(\mathbf{A},r)$ indicates top $r$ singular vectors of matrix $\mathbf{A}$, and $\text{eigs}(\mathbf{A},r)$ indicates top $r$ eigenvectors of matrix $\mathbf{A}$. The detailed derivation is listed in the Appendix. This iterative procedure is termed as Joint HOOI Algorithm presented in Algorithm~\ref{alg2}.
\begin{algorithm}[H]
\caption{Joint HOOI Algorithm for minimizing~(\ref{M2_sub2}).}
\label{alg2}
\begin{algorithmic}[1]
\REQUIRE The initialization of $\mathbf{U}_{1p}$, $\mathbf{U}_{2p}$, $\mathbf{U}_3$ and $\mathbf{U}_{4p}$; $\mathcal{R}_p(\widetilde{\mathcal{X}_1})$.
\ENSURE $\mathbf{U}_{1p}$, $\mathbf{U}_{2p}$, $\mathbf{U}_3$ and $\mathbf{U}_{4p}$; $\mathcal{G}_p$.
\WHILE{ not converged }
\STATE Updating $\mathcal{G}_p$ via Eq.~(\ref{alg2:eq1});
\STATE Updating $\mathbf{U}_{1p}$, $\mathbf{U}_{2p}$ and $\mathbf{U}_{4p}$ via Eqs.~(\ref{alg2:eq2}),~(\ref{alg2:eq3}) and~(\ref{alg2:eq4});
\STATE Updating $\mathbf{U}_3$ via Eq.~(\ref{alg2:eq5}).
\ENDWHILE
\end{algorithmic}
\end{algorithm}
The algorithm for the PG-TenRPCA model now can be easily designed through replacing step 3 in~\textbf{Algorithm}~\ref{alg1} by solving the optimization problem in Eq.~\eqref{M2_sub2}.
Additionally, the clustering, or say the updating of $\mathbf{R}_p~(p=1,2,\cdots,K)$ is performed every some iterations, e.g., 8 iterations, and the first clustering is performed over an initialized video background.

It is obvious that our proposed models are non-convex and non-separable optimization problems. Therefore, there may exist many local minimizers and a suitable initialization is crucial for attaining the desired solution. Although the convergence of the multi-block ADMM for this kind of optimization problems, to the best of our knowledge, is not guaranteed, the experimental results in Section VI will justify that given a suitable initialization, the proposed algorithms based on the multi-block ADMM with the adaptive scheme can produce satisfactory results. Specifically, for the optimization algorithm of H-TenRPCA, we initialize $\mathcal{L}$ by $(r_1,r_2,r_3)$-Tucker decomposition of $\textbf{Ten}(\mathcal{A}^{*}(\mathbf{y}))$ and $\mathbf{x}_2$ by $\mathcal{A}^{*}(\mathbf{y})$-$\textbf{Vec}(\mathcal{L})$. For optimization algorithm of PG-TenRPCA, the result from H-TenPCA algorithm provides a suitable initialization.
\subsection{Implementation Issues}
In~\textbf{Algorithm}~\ref{alg1}, there exist four parameters, i.e., $r_1$, $r_2$, $r_3$ and $\lambda$, where $r_1$ and $r_2$ control the complexity of spatial redundancy, $r_3$ controls the complexity of temporal redundancy, and $\lambda$ provides a trade-off between disturbance and foreground modeling. $r_1$ and $r_2$ for factor matrices $\mathbf{U}_1$ and $\mathbf{U}_2$ are empirically taken as $r_1=\text{ceil}$\footnote{\text{ceil}($a$) indicates the smallest integer larger than $a$.}$(H\times 0.65)$ and $r_2=\text{ceil}(W\times 0.65)$ in all conducted experiments and we indeed find this setting works fairly well. Actually, such selected $r_1$ and $r_2$ can make the AccEgyR index attain the ratio over 0.9 for various natural images and we have showed some examples in Fig. 2(d).
For $r_3$ and $\lambda$, it is required to carefully tune them for testing data sets. We empirically found that our algorithm will achieve satisfactory performance when $r_3$ is taken as the value 1 for the real-world data sets and $\lambda$ is taken in the range $\lbrack 0.01,0.1\rbrack$.

In the optimization algorithm for PG-TenRPCA model in Eq.~\eqref{rs_model2}, we need to set nine parameters, i.e., the size of 3D patch $w$, the size of search window around one patch $S$, the number of collected similar 3D patches $N$, the sliding distance $d$, the rank parameters $r_1$, $r_2$, $r_3$, $r_4$, and the traded-off parameter $\lambda$. Empirically, $w$, $d$, $S$ and $N$ are respectively taken as 8, 7, 36, and 45~\cite{Dong_b,Dong_d}. The rank constraint parameters $r_1$, $r_2$, $r_4$ are respectively set to 8, 8, and $\text{ceil}(45\times 0.45)$. Here $r_4=\text{ceil}(45\times 0.45)$ implies that our algorithm makes low rank approximation due to the large redundancy hidden in the similar patches. Similar to~\textbf{Algorithm}~\ref{alg1}, we only need to carefully tune $r_3$ for temporal complexity and the trade-off parameter $\lambda$. We empirically found that $r_4$ is taken as the value 1 for the real world data sets, and $\lambda$ is taken in the range $\lbrack 0.05, 0.1 \rbrack$.
It is worth noting that in order to reduce the computational cost, the clustering is performed by the K-nearest neighbor method. Specifically, for each 3D patch of the patch set~$\mathcal{S}$, we search N similar 3D patches as a cluster from a big window around this 3D patch.
\section{Experimental Results}
In this section, we will conduct experiments on synthetic and real video datasets to demonstrate the superiority of two proposed models, i.e., H-TenRPCA and PG-TenPCA,  over the existing state-of-the-art approaches for the BSCM task. All the experiments are performed using MATLAB (R2013a) on workstations with dual-core Intel processor of 2.90 GHz and RAM of 30 GB equipped with Windows 7 OS.
The parameter tuning is performed by grid search for our proposed methods as well as the compared methods such that the following \textit{averaged} PSNR index over video frames achieves the best value.

We first introduce the evaluation measures. We use $\text{F-measure}$ to assess the detection performance of video foreground, and the peak signal-to-noise ratio (PSNR) and the structural similarity index (SSIM) to measure the reconstruction accuracies. $\text{F-measure}$ is defined as:
$\text{F-measure}=2\frac{\text{precision}\cdot \text{recall}}{\text{precision+recall}},$
where \text{recall} and \text{precision} are defined as:
\begin{displaymath}
\text{recall}=\frac{\text{\#correctly classified foreground pixels}}{\text{\#foreground pixels in ground truth}},
\end{displaymath}
\begin{displaymath}
\quad\quad \text{precision}=\frac{\text{\#correctly classified foreground pixels}}{\text{\#pixels classified as foreground}}.
\end{displaymath}
PSNR and SSIM commonly measure
the similarity of two images in intensity and structure respectively.~PSNR is defined as:
$\text{PSNR}: = 10\times \text{log}_{10}\frac{255^2}{\sum_{ij}(I_{ij} - \hat{I}_{ij})^2},$
where $I_{ij}$ and $\hat{I}_{ij}$ are respectively the intensity values of the original and reconstruction images at the pixel $(i,j)$.
SSIM measures the structural similarity of two images; see~\cite{SSIM} for details. We use \textit{averaged} PSNR and SSIM over video frames to evaluate reconstruction performance of video volume. Higher values of F-measure, PSNR and SSIM indicate the better performance.
\begin{figure}[htp]
\centering\vspace{-2mm}
\includegraphics[width=3.3in,height=2.45in]{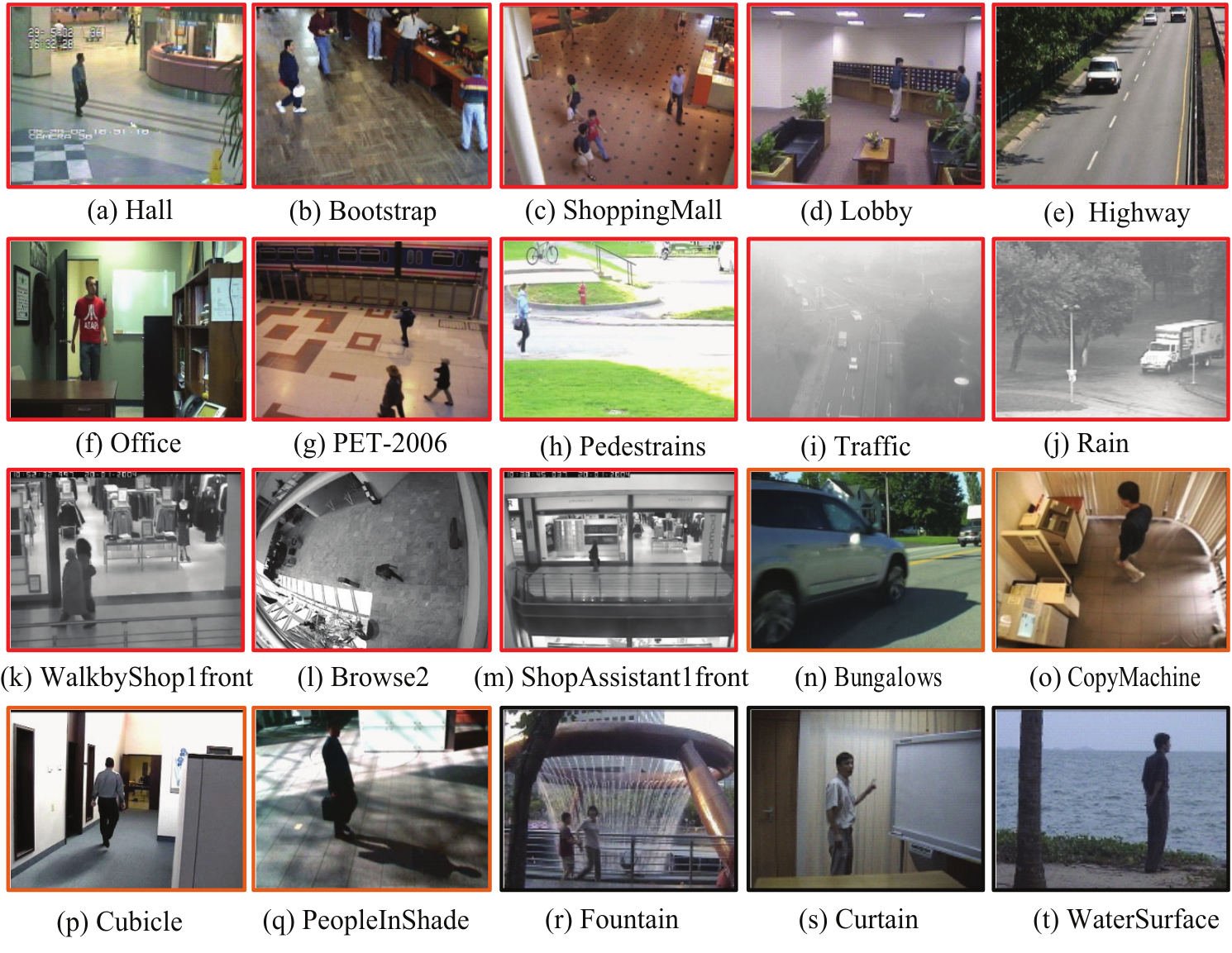}
\vspace{-2mm}
\caption{Sampled images from real videos.}\vspace{-4mm}
\label{realData}
\end{figure}
\subsection{Data Sets}
\subsubsection{Synthetic Data}
The SABS\footnote{http://www.vis.uni-stuttgart.de/index.php?id=sabs} (Stuttgart Artificial Background Subtraction) dataset is an artificial dataset for pixel-wise evaluation of background models. The dataset consists of video sequences for nine different challenges of background subtraction. The \textit{basic} class of nine different challenges is used to evaluate our proposed approach. We collect 128 frames (say, NoForegroundDay0001$\rightarrow$NoForegroundDay0128) from the SABS-basic data, and then scale each frame into an image of size 128$\times$128 as a frame of the true background. Similarly, we choose 128 frames (say, GT0807-GT0934) as the foreground from SABS-GT data and then transform the intensity of these gray images into the range from 200 to 255 for visual contrast to the background. Then, it is easy to obtain the original video volume $\mathcal{X}_0$ by combining the background $\mathcal{X}_1$ and the foreground $\mathcal{X}_2$.
The example video shown in Fig.~\ref{mdlPri} is from this dataset.
\subsubsection{Real Data}
We collect a set of real world videos from CAVIAR dataset\cite{data2003}\footnote{http://groups.inf.ed.ac.uk/vision/CAVIAR/CAVIARDATA1/}, I2R dataset~\cite{Li2004}\footnote{http://perception.i2r.a-star.edu.sg/bk\underline{  }model/bk\underline{  }index.html}, UCSD dataset~\cite{data2010}\footnote{http://www.svcl.ucsd.edu/projects/background\underline{  }subtraction/} and CD.net dataset~\cite{Goyette2012}\footnote{http://changedetection.net}.
These data sets include various real world scenes ranging from the simple scenes with~static backgrounds to the complex scenes with camera jitter or intermittent object motion. From these data sets, we choose three categories of videos for testing our approach:~\textit{static background} (Fig.~\ref{realData}(a)-(m)), \textit{shadow} (Fig.~\ref{realData}(n)-(q)), and \textit{dynamic background} (Fig.~\ref{realData}(r)-(t)). For each video, 128 gray-scale video frames are chosen as video volume for our experiments.
\begin{figure*}[htp]
\centering\vspace{-4mm}
\includegraphics[width=5.85in,height=2.7in]{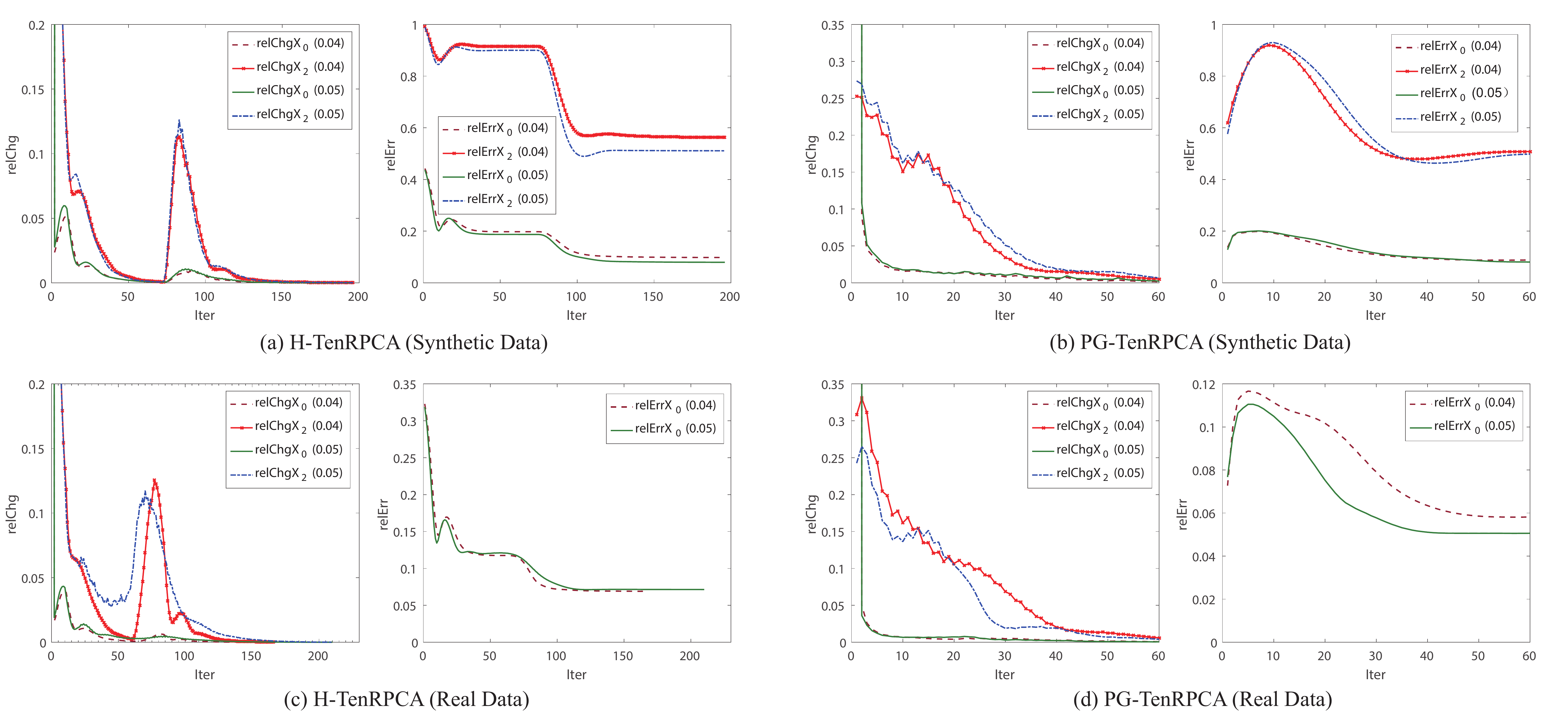}
\vspace{-1mm}
\caption{The empirical analysis of algorithm convergence}
\label{Converg}
\end{figure*}
\subsection{Empirical Analysis for Algorithm Convergence}
We provide an empirical analysis for the convergence of the proposed optimization algorithms on a synthetic video shown in Fig.~\ref{mdlPri}, and a real video ``ShoppingMall'' shown in Fig.~\ref{realData}(c). The relative change $\text{relChg}\mathbf{A}:= \frac{\|\mathbf{A}^{k} - \mathbf{A}^{k-1} \|_\text{F}} { \text{max}(1, ~\|\mathbf{A}^{k-1} \|_\text{F})}$ and the relative error $\text{relErr}\mathbf{A}:= \frac{\|\mathbf{A}^{k} - \mathbf{A}_{0} \|_\text{F}} { \text{max}(1, ~\|\mathbf{A}_{0} \|_\text{F})}$ are used as the assessment index of algorithm convergence, where $\mathbf{A}^{k}$ is the result in $k$-th iteration and $\mathbf{A}_{0}$ is the ground-truth result.

In Fig.~\ref{Converg}, we show the curves of  the relative change and the relative error of video volume $\mathbf{X}_0$ and video foreground $\mathbf{X}_2$  for algorithms H-TenRPCA and PG-TenRPCA, where the sampling ratio is set as 0.04 (1/25) and 0.05 (1/20), respectively. $\text{relChg}\mathbf{X}_2$ denotes the relative change of video foreground $\mathbf{X}_2$. In Fig.~\ref{Converg}(a)-(b), we show the convergence results of H-TenRPCA and PG-TenRPCA on the synthetic video, and in Fig.~\ref{Converg}(c)-(d),
the convergence results on a real video. Note that in Fig.~\ref{Converg}(c)-(d), we do not provide the convergence results for the relative error of video foreground, because the ground-truth video foreground for real data is unknown.

Generally, the relative change converges to zero when the number of iterations is high, and the corresponding relative error w.r.t. ground-truth
gradually decreases to a stable value. From Fig.~\ref{Converg}(a) and~(c), we observe a significant jump of relative change for video foreground $\text{relChg}\mathbf{X}_2$ between iterations [60, 100], and the jump corresponds to a large decrease of relative error shown in the right subfigures of Fig.~\ref{Converg}(a) and (c). Thus this jump corresponds to a sudden significant improvement on the foreground estimation during the optimization procedures.
From all subfigures in Fig.~\ref{Converg}, we observe that the curves of all assessment indices reduce to a stable value when the algorithms reach a relatively high iteration number, which suggests that the proposed algorithms  well converge empirically.

\subsection{Comparison with Existing Popular Methods}
We compare our models H-TenRPCA and PG-TenRPCA with three existing popular methods: SpaRCS~\cite{Waters2011}, SpLR~\cite{Jiang2012a} and ReProCS~\cite{Guo2014}.
The SpaRCS and SpLR methods are both batch-based approaches as ours that process a batch of video frames, i.e., a video volume, as a whole. Whereas, ReProCS is an online method that processes the video frames sequentially. It requires to use the training video frames to initialize a video background, and requires the compressive operator over each frame to be the same. For fair comparison, we thus create an additional subsection to compare with the ReProCS method.
\subsubsection{Comparison with Batch-Based Methods}
In this subsection, we compare our approach with SpaRCS~\cite{Waters2011} and SpLR~\cite{Jiang2012a}  on synthetic data and real data sets. Considering the feasibility on current chips of CS cameras, the randomly permuted Walsh-Hardmard in the frame-wise manner is chosen as compressive operator.  That is, the compressive operator is chosen as $\mathbf{D}_d \cdot \mathbf{H}_d \cdot \mathbf{P}_d~(d=1,\cdots,D)$ for all compared methods. The sampling ratios are set as two high levels 1/5 and 1/10, and three low levels 1/20, 1/25 and 1/30 for assessing
the reconstruction and separation performance of all compared methods. For illustrating the merits of tensor modeling technique, we also compare the degenerated version of our method H-TenRPCA, where video background is modeled by a low rank matrix instead of a tensor on the synthetic video data. The degenerated version is dubbed as H-MatRPCA.
\begin{table}[htp]
\centering
\scriptsize
\setlength{\tabcolsep}{3pt}
\caption{Comparison of different methods on the synthetic video. Note that PSNR here indicates the averaged PSNR on all video frames. The same for SSIM and F-measure.}
\begin{tabular}{ccccccc}
\toprule[1pt]
SR & Indices &  SpaRCS & SpLR & H-MatRPCA & H-TenRPCA & PG-TenRPCA \\ \hline
\multirow{3}{*}{1/5}   & PSNR &   26.42    &  45.07   & \textbf{45.38} & 42.33    &   40.15  \\
                       & SSIM &  0.8678      & 0.9955 & \textbf{0.9963} & 0.9918&  0.9894  \\
                       & F-measure &  0.6194    & \textbf{0.9195}&  0.8617  &  0.8624 & 0.8598   \\
\hline
\multirow{3}{*}{1/10}  & PSNR &   17.03  & 34.09 & \textbf{35.16} & 34.95 &  34.38   \\
                       & SSIM &  0.4723     & 0.9627   &   \textbf{0.9726} & 0.9695 &  0.9673  \\
                       & F-measure &  0.0704     & \textbf{0.8909}  & 0.8601 &0.8616 &   0.8566   \\  \hline
\multirow{3}{*}{1/20}   & PSNR & 14.14  & 25.40  &   30.53 & \textbf{30.64}&     30.40  \\
                       & SSIM &  0.2405   &  0.8184  &  0.9245 & 0.9275 &  \textbf{0.9299} \\
                       & F-measure &  0.0333 & 0.7069 &    0.8327  & 0.8337   & \textbf{0.8342} \\  \hline
\multirow{3}{*}{1/25}   & PSNR &   13.86  &  23.57   &   28.45     &   28.80&    \textbf{29.61} \\
                       & SSIM &  0.2129   &    0.7541    &    0.8776   &    0.8901   &     \textbf{0.9236} \\
                       & F-measure & 0.0311  &     0.5854  &      0.8172    &    0.8182  &    \textbf{0.8240} \\  \hline
\multirow{3}{*}{1/30}   & PSNR &  13.52   & 22.46   &   26.92   &    27.36 &   \textbf{28.88}  \\
                       & SSIM &   0.1782    & 0.7039   &   0.8268    &     0.8492     &      \textbf{0.9133}  \\
                       & F-measure &   0.0301  &   0.4767   &   0.8080      &     0.8086   &      \textbf{0.8148}
  \\
\bottomrule[1pt]
\end{tabular}
\label{simuTab1}
\end{table}
\begin{figure*}[htp]
\centering\vspace{-5mm}
\includegraphics[width=6.3in,height=3.35in]{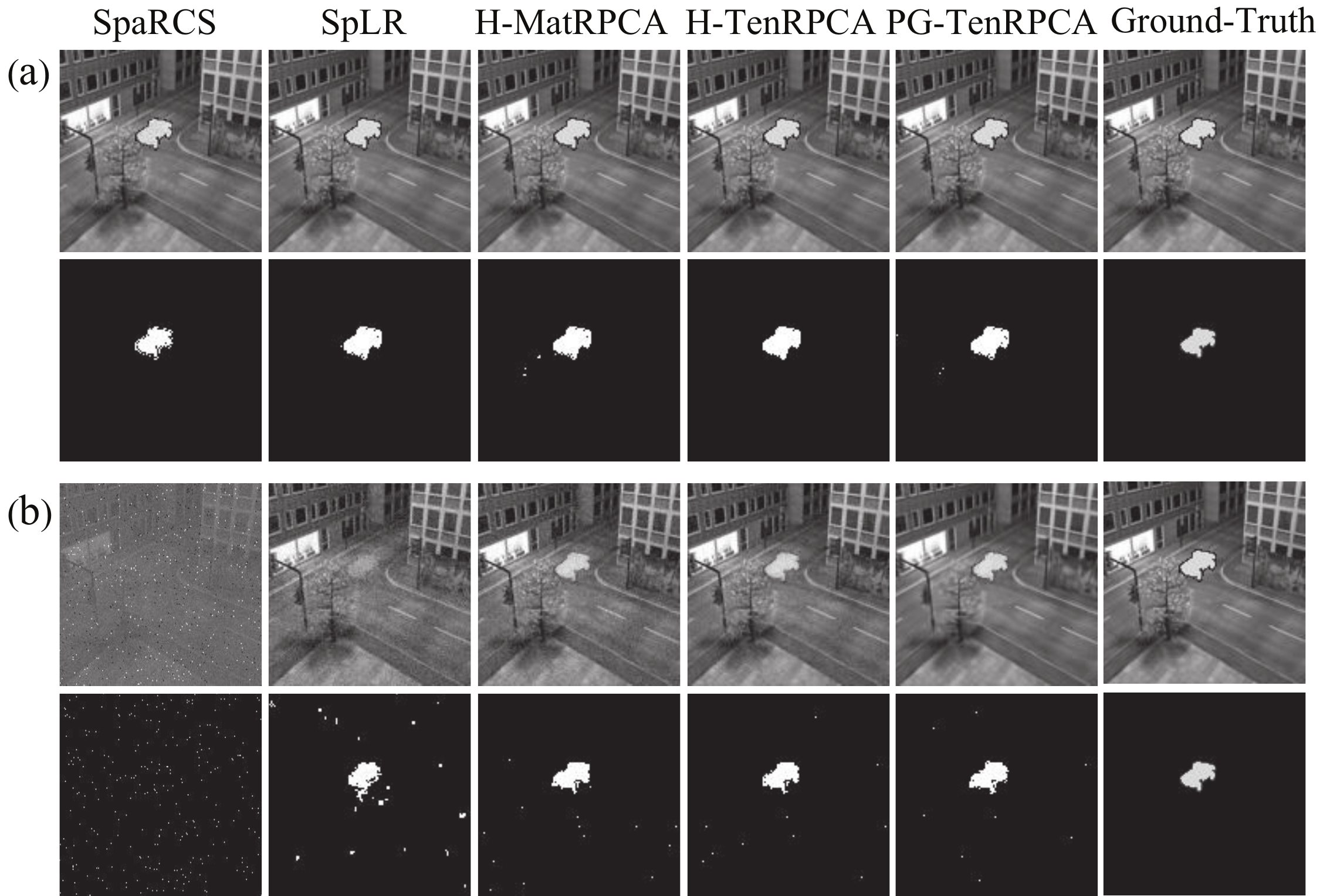}
\caption{Comparison of visual results of different methods on a frame of the synthetic video. (a) shows the results of reconstruction and detection of one video frame with a high sampling ratio 0.2 (1/5); while (b) shows the case with a low sampling ratio 0.04 (1/25) for comparison.}\vspace{-2mm}
\label{simuWHTfig}
\end{figure*}

We show the quantitative results of all compared methods with different sampling ratios on the synthetic data in Table~\ref{simuTab1}. The averaged PSNR and SSIM values indicate the reconstruction performance of the original video, and the averaged F-measure values indicate the separation
(or detection) performance of video foreground. We observe that, when the sampling ratio is taken as a high value of 1/5, all methods can reconstruct the original video and detect a satisfactory silhouette of video foreground. When the sampling ratio goes down, our proposed models perform consistently better than all the compared methods. First, our tensor based models work significantly better than the conventional methods, i.e., SpaRCS and SpLR, and also the matrix version of our model, i.e., H-MatRPCA.  Second, the PG-TenRPCA model that is based on video patch groups works better than the H-TenRPCA model that takes the video volume as a single tensor.

Figure~\ref{simuWHTfig}(a) and (b) show
the visual results when the sampling ratio (SR) is taken as 1/5 and 1/25 respectively. The last column shows the ground-truth videos and silhouettes of video foregrounds.  It can be observed that, when the sampling ratio is a low value of 1/25, the SpaRCS method totally fails in reconstructing the original video and detecting the moving car. Although the SpLR method can produce a slightly better result, the car in the reconstructed video is blurred and the detected car is incomplete and disturbed by noisy points. Compared with the ground-truth in the last column, H-MatRPCA, H-TenRPCA and PG-TenRPCA can produce satisfactory visual results but the reconstructed video by H-MatRPCA is not clear and sharp as two other models. It is worth noting that,~compared with other methods, the reconstructed video by PG-TenRPCA is very clear and sharp due to the powerfulness of nonlocal self-similarity prior.

\begin{figure}[htp]\vspace{-2.5mm}
\centering
\includegraphics[width=3.4in,height=2.4in]{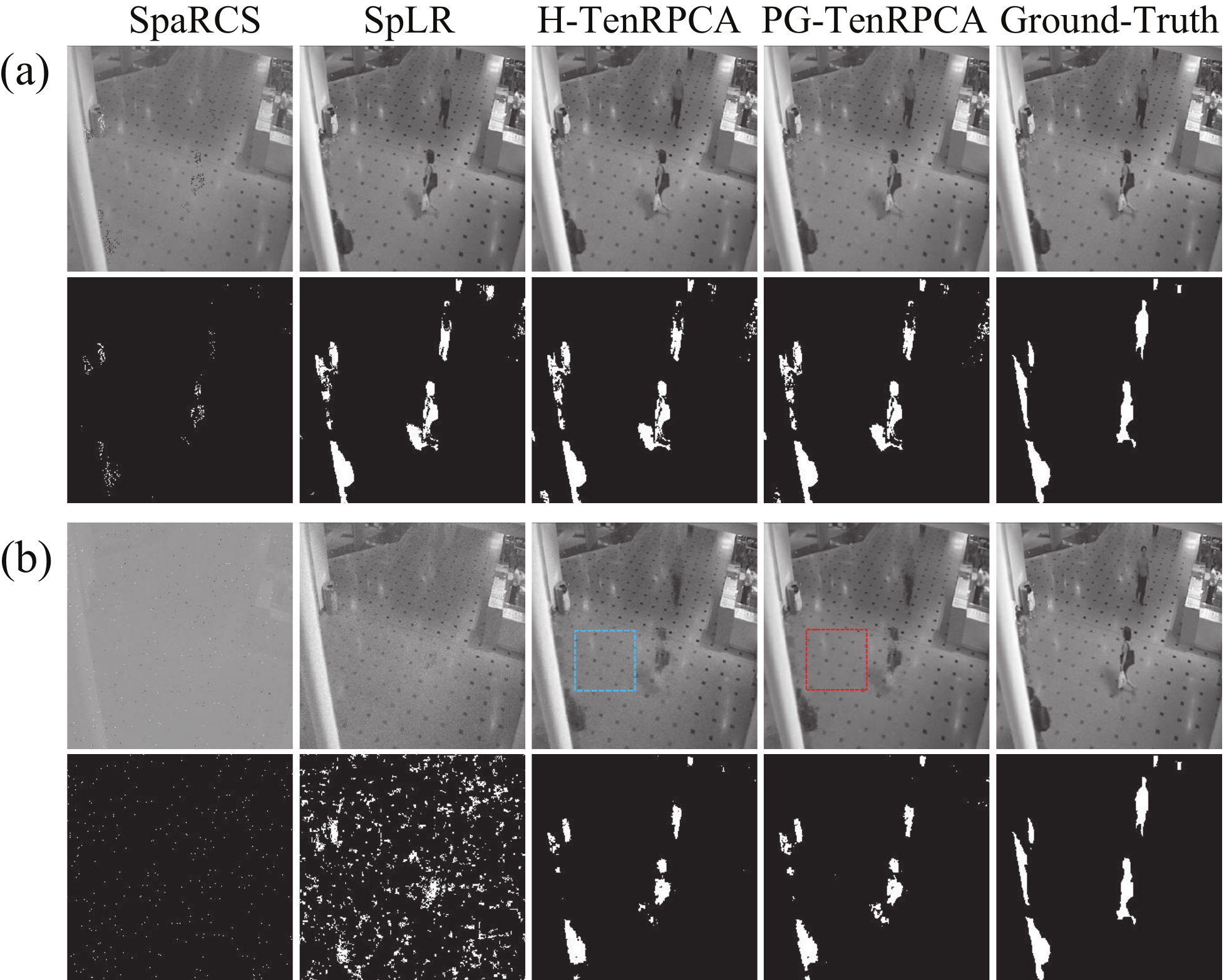}
\caption{Visual results of different methods on a frame of the ShoppingMall video. (a) shows the results of reconstruction and detection of one video frame with a high sampling ratio 0.2 (1/5); while (b) shows the case with a low sampling ratio 0.04 (1/25) for comparison.}
\label{realWHTfig}
\end{figure}

\begin{figure}[htp]
\centering\vspace{-3mm}
\includegraphics[width=3.25in,height=1.42in]{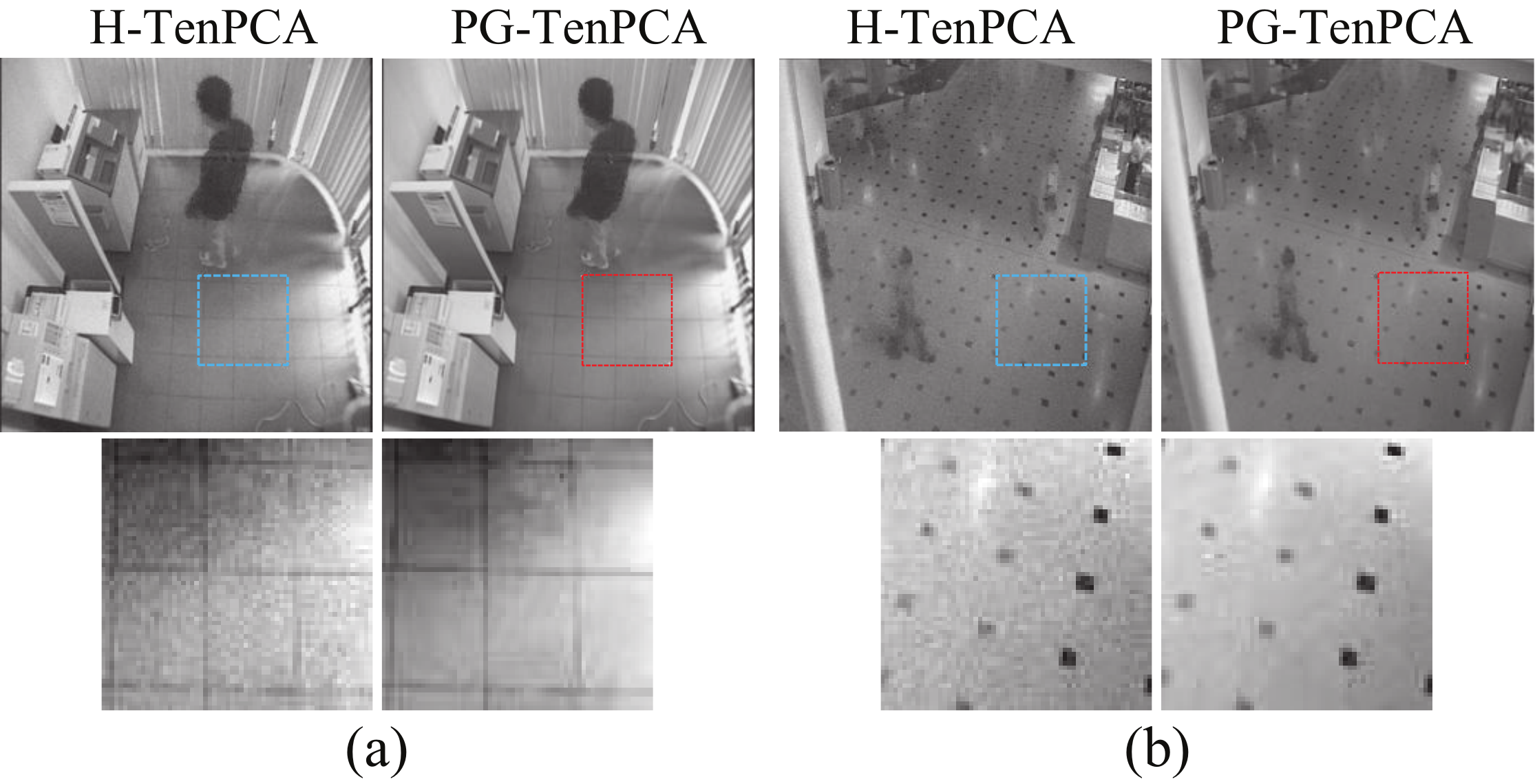}
\centering\vspace{-1mm}
\caption{Comparison of PG-TenRPCA with H-TenRPCA with respect to reconstruction performance with the sampling ratio 0.04.}
\label{NLgood1}
\end{figure}

\begin{figure}[htp]
\centering
\includegraphics[width=3.4in,height=6.25in]{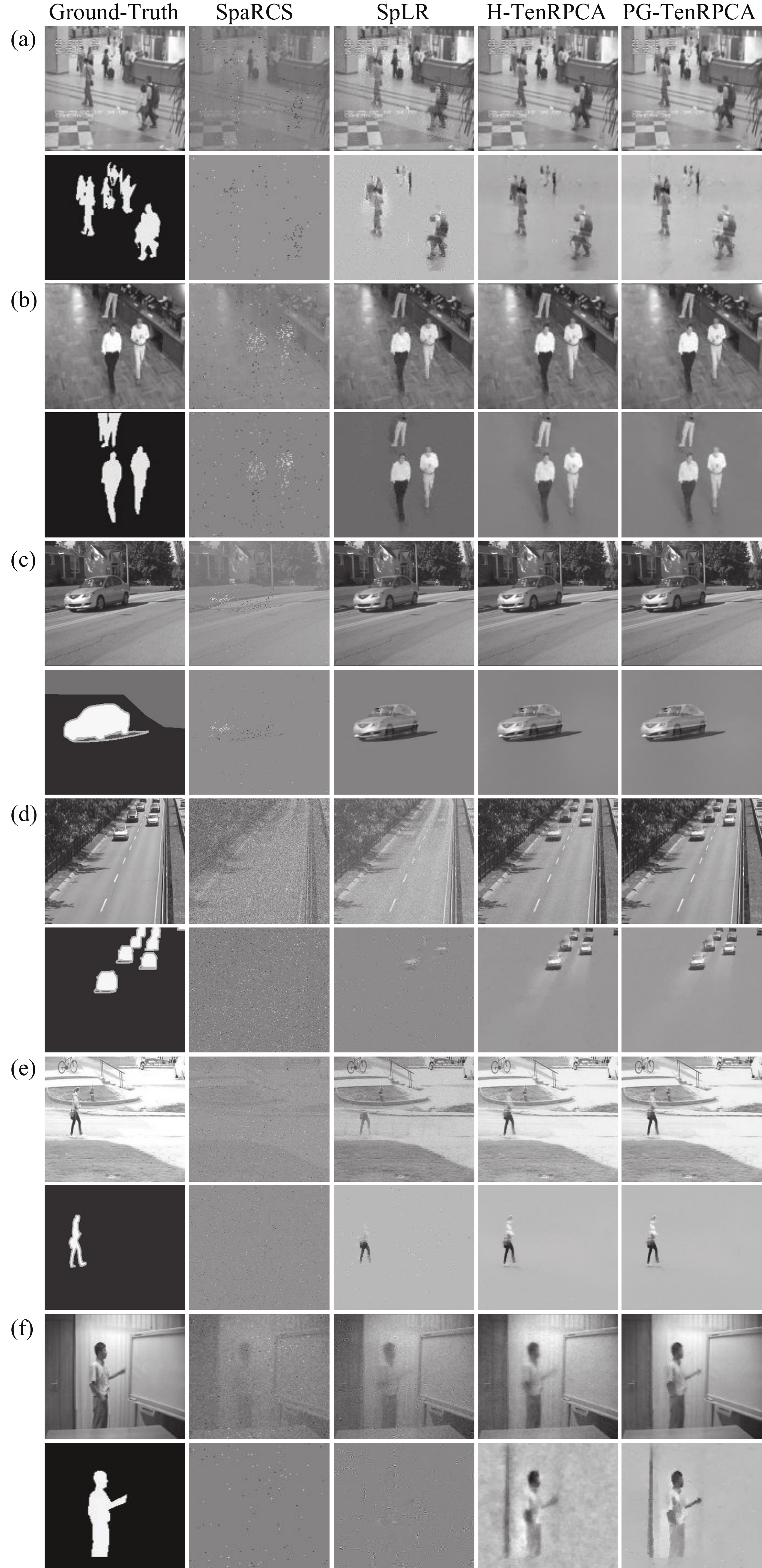}
\caption{The reconstruction and separation (RS) results of different methods on six videos. The first three videos show the RS results for a high sampling ratio 0.2, while the last three videos for a low sampling ratio 0.04.}
\label{realDataShow} \vspace{-5mm}
\end{figure}

We also present the quantitative results of all compared methods on a real captured video ``ShoppingMall'' with labeled foregrounds in several video frames in Table~\ref{realTab1}. The visual comparison results on this video are shown in Fig.~\ref{realWHTfig}. Compared with the synthetic video, this video is more challenging due to the multiple walking persons in the video. From the Table~\ref{realTab1} and Fig.~\ref{realWHTfig}, we can observe that, when the sampling ratio is taken as a high value of 1/5,
all methods except the SpaRCS method can reconstruct a high-quality video and detect relatively satisfactory silhouettes of the walking persons. The SpaRCS method failed in detecting the walking persons in Fig.~\ref{realWHTfig}(a), which might be because of the insufficiency of the simple sparse prior used for video foreground in SpaRCS. When the sampling ratio goes down, the SpLR method also failed in detecting moving objects. However, our proposed H-TenRPCA and PG-TenRPCA models can still produce satisfactory results.  Moreover, compared with the H-TenRPCA method, the PG-TenRPCA method works better both visually and quantitatively, because it uses a well-designed patch-based prior, i.e., the nonlocal self-similarity, to model the patch-level correlations of video background.  As shown in Fig.~\ref{realWHTfig}(b),
the reconstructed region indicated by the red box is more clear and sharper than the region indicated by the light blue box (Best seen in the zoom-in version of pdf), which can be further illustrated in Fig.~\ref{NLgood1}.
\begin{table}[htp]
\centering
\scriptsize
\caption{Comparison of different methods on the ShoppingMall video. Note that PSNR here indicates the averaged PSNR on all frames. The same for SSIM and F-measure.}
\begin{tabular}{ccccccc}
\toprule[1pt]
SR & Indices &  SpaRCS & SpLR  & H-TenRPCA & PG-TenRPCA \\ \hline
\multirow{3}{*}{1/5}  & PSNR &  25.55     & 35.38 & \textbf{41.02}  & 40.25 \\
                       & SSIM &  0.8290     & 0.9468  & \textbf{0.9768} & 0.9736  \\
                       & F-measure &  0.1086    & 0.6647 & \textbf{0.6714} & 0.6672  \\  \hline
\multirow{3}{*}{1/10}   & PSNR &  24.41    & 28.85 & \textbf{37.03} & 36.38    \\
                       & SSIM &  0.7462     &  0.8434 & \textbf{0.9574 }& 0.9535  \\
                       & F-measure &  0.0366   & 0.5318  & \textbf{0.6611}

 &  0.6568   \\
\hline

\multirow{3}{*}{1/20}   & PSNR &  22.37     & 25.19 & 31.34 & \textbf{32.48}  \\
                       & SSIM &  0.5730    & 0.6876 & 0.8779  & \textbf{0.9197}   \\
                       & F-measure & 0.0109     & 0.2207 & 0.6186 & \textbf{0.6314}   \\  \hline
\multirow{3}{*}{1/25}   & PSNR &  21.49     & 24.56  & 29.85  & \textbf{31.47}   \\
                       & SSIM &  0.5155    &  0.6378 &  0.8259  & \textbf{0.9058 }  \\
                       & F-measure &  0.0123    & 0.1646  & 0.6008 & \textbf{0.6146 }  \\  \hline
\multirow{3}{*}{1/30}   & PSNR &  20.76    & 23.97 & 28.38  & \textbf{30.71}  \\
                       & SSIM & 0.4651    & 0.5847 & 0.7561 & \textbf{0.8948}    \\
                       & F-measure &  0.0099    & 0.1298 & 0.5842 & \textbf{0.5964}   \\
\bottomrule[1pt]
\end{tabular}
\label{realTab1}
\end{table}
\begin{table}[htp]
\centering
\scriptsize
\caption{The comparison results on real data with sampling ratio 1/25. Note that PSNR here indicates the averaged PSNR on all frames. The same for SSIM and F-measure. We ignore the statistics of F-measure by the notation ``-". Because the ground-truth silhouette of video foreground is not provided.}
\begin{tabular}{cccccc}
\toprule[1pt]
Videos & \multicolumn{1}{c}{Indices} &\multicolumn{1}{c}{SpaRCS}  & \multicolumn{1}{c}{SpLR} & \multicolumn{1}{c}{H-TenRPCA} & \multicolumn{1}{c}{PG-TenRPCA} \\ \hline
\multirow{3}{*}{a} & PSNR &17.85   &20.58  & 27.28  & $\textbf{28.48}$    \\ 
                      & SSIM  &0.4974 & 0.6310 &0.8469 & $\textbf{0.9077}$  \\ 
                      & F-measure  &0.0049  &0.2534 &0.6041 & $\textbf{0.6132}$ \\ \hline
\multirow{3}{*}{b} & PSNR &18.51   &21.88  & 27.11  & $\textbf{28.91}$    \\ 
                      & SSIM  &0.4226 & 0.5819 &0.7929& $\textbf{0.8786}$  \\ 
                      & F-measure  &0.0035  &0.2577 &0.6404 & $\textbf{0.6707}$ \\ \hline
\multirow{3}{*}{c} & PSNR &21.49   &24.56  & 29.85  & $\textbf{31.47}$    \\ 
                      & SSIM  &0.5155 & 0.6378 &0.8259 & $\textbf{0.9058}$  \\ 
                      & F-measure  &0.0043&0.1646 &0.6008  & $\textbf{0.6146 }$ \\ \hline
\multirow{3}{*}{d} & PSNR &25.83   &34.13 & \textbf{40.92}& 36.90    \\ 
                      & SSIM  &0.7010 & 0.9029 &\textbf{0.9792} & 0.9737  \\ 
                      & F-measure  &0.0131  &0.5505 & \textbf{0.6938}& 0.6584 \\ \hline
\multirow{3}{*}{e} & PSNR &13.97   &20.27  & 27.59  & $\textbf{29.47}$    \\ 
                      & SSIM  &0.1971 &0.5109&0.7949 & $\textbf{0.8980}$  \\ 
                      & F-measure  &0.0749   &0.1794&0.6218  & $\textbf{0.6444}$ \\ \hline
\multirow{3}{*}{f} & PSNR &10.38  &21.39 & 28.40  & $\textbf{34.41}$    \\ 
                      & SSIM  &0.1077 & 0.5167 &0.8247 & $\textbf{0.9565}$  \\ 
                      & F-measure  &0.0479  &0.1156&0.6203 & $\textbf{0.6785}$ \\ \hline
\multirow{3}{*}{g} & PSNR &13.26   &25.45  & 38.52 & $\textbf{40.10}$    \\ 
                      & SSIM  &0.1678 & 0.6823 &0.9603& $\textbf{0.9794}$  \\ 
                      & F-measure  &0.0320  &0.5937&0.7663 & $\textbf{0.7708}$ \\ \hline
\multirow{3}{*}{h} & PSNR &16.25   &26.82  & 35.99  & $\textbf{36.57}$    \\ 
                      & SSIM  &0.3474 & 0.7490 &0.9472 & $\textbf{0.9661}$  \\ 
                      & F-measure  &0.0171  &0.5245 &0.6086 & $\textbf{0.6141}$ \\ \hline
\multirow{3}{*}{i} & PSNR &19.27   &35.73  & 39.98  & $\textbf{43.23}$    \\ 
                      & SSIM  &0.4205 & 0.8627 &0.9459& $\textbf{0.9670}$  \\ 
                      & F-measure  &0.0238  &0.1078 &0.2651 & $\textbf{0.3478}$ \\ \hline
\multirow{3}{*}{j} & PSNR &19.46  &33.06  & 36.11  & $\textbf{38.15}$    \\ 
                      & SSIM  &0.4138 & 0.8288 &0.9076 & $\textbf{0.9378}$  \\ 
                      & F-measure  &-  &-&- & - \\ \hline
\multirow{3}{*}{k} & PSNR &19.73   &23.55  & 32.09  & $\textbf{37.88}$    \\ 
                      & SSIM  &0.5361 & 0.6545 &0.9125 & $\textbf{0.9771}$  \\ 
                      & F-measure  &-  &- &-& - \\ \hline
\multirow{3}{*}{l} & PSNR &24.03   &16.36  & 31.15 & $\textbf{37.05}$    \\ 
                      & SSIM  & 0.6235 & 0.4018 &0.8417 & $\textbf{0.9612}$  \\ 
                      & F-measure  &-  &- &- & - \\ \hline
\multirow{3}{*}{m} & PSNR &25.24   &31.03  & 36.80  & $\textbf{38.90}$    \\ 
                      & SSIM  &0.7183 & 0.8715 &0.9725 & $\textbf{0.9873}$  \\ 
                      & F-measure  &-  &-&- & - \\ \hline
\multirow{3}{*}{n} & PSNR &16.70   &21.29 & 28.32  & $\textbf{31.22}$    \\ 
                      & SSIM  &0.3441 &0.4603 &0.7510 & $\textbf{0.8593}$  \\ 
                      & F-measure  & 0.0022  &0.1654&0.3182& $\textbf{0.3249}$ \\ \hline
\multirow{3}{*}{o} & PSNR &15.57  &19.24 & 31.83  & $\textbf{33.83}$    \\ 
                      & SSIM  &0.3304 & 0.4321 & 0.8726& $\textbf{0.9393}$  \\ 
                      & F-measure  &0.0038  &0.1556 &0.8210 & $\textbf{0.8307}$ \\ \hline
\multirow{3}{*}{p} & PSNR &15.18   &20.02  & 27.65  & $\textbf{34.79}$    \\ 
                      & SSIM  &0.2733 & 0.4052 &0.7610& $\textbf{0.9454}$  \\ 
                      & F-measure  &0.0028 &0.1159 &0.5534 & $\textbf{0.6128}$ \\ \hline
\multirow{3}{*}{q} & PSNR &17.67   &21.45 & 27.19  & $\textbf{33.77}$    \\ 
                      & SSIM  &0.3648 & 0.5137 &0.7899 & $\textbf{0.9466}$  \\ 
                      & F-measure  &0.0035 &0.1915 &0.7056 & $\textbf{0.7812}$ \\ \hline
\multirow{3}{*}{r} & PSNR &21.78   &26.68 & \textbf{30.78 } & 30.64   \\ 
                      & SSIM  &0.6817 & 0.8338 & 0.9136 & $\textbf{0.9257}$  \\ 
                      & F-measure  &0.0064 &0.2954& 0.7450 & $\textbf{0.7602}$ \\ \hline
\multirow{3}{*}{s} & PSNR &21.03   &23.92 & 28.93  & $\textbf{32.72}$    \\ 
                      & SSIM  &0.4253& 0.5449 & 0.7643 & $\textbf{0.8975}$  \\ 
                      & F-measure  &0.0025&0.1726 &0.5257& $\textbf{0.5845}$ \\ \hline
\multirow{3}{*}{t} & PSNR &17.57   &22.20  & 29.79  & $\textbf{30.67}$    \\ 
                      & SSIM  &0.3183 & 0.4678 &0.7450 & $\textbf{0.7912}$  \\ 
                      & F-measure  &0.0026 & 0.1795 & 0.8515 & $\textbf{0.8534
}$ \\
\bottomrule[1pt]
\end{tabular}
\label{realTab2}
\end{table}
We further provide more experimental results on various real videos to demonstrate the effectiveness of our proposed models, especially for the low sampling ratios. In Table~\ref{realTab2} and Table~\ref{realTab3}, we show the quantitative results on multiple real videos with sampling ratio of 1/25 and the averaged quantitative results on these videos with different sample ratios respectively.
Observed from Table~\ref{realTab2}, our tensor-based models perform significantly better with much higher PSNR,~SSIM and F-measure values than all the compared methods at a low sampling ratio of 1/25. From Table~\ref{realTab3}, we observe that, our proposed models work overall better across different sampling ratios. At a high sampling ratio of 1/5, the compared methods of SpaRCS and SpLR can also reconstruct the original video and detect foregrounds with gracefully high values of PSNR, SSIM and F-measure, but still significantly lower than ours. When the sampling ratio goes down, the models SpaRCS and SpLR fail to well reconstruct the videos and separate the video foregrounds, but our proposed models of H-TenRPCA and PG-TenRPCA can still perform very well with high values of PSNR, SSIM and F-measure. Moreover, the PG-TenRPCA model is superior over H-TenRPCA on average. In Fig.~\ref{realDataShow}, we further visually show the results of reconstruction and separation by different methods on real six videos. Figure~\ref{realDataShow}(a)-(c) show results of three videos with a high sampling ratio of 1/5,
and Figure~\ref{realDataShow}(d)-(f) show results of three videos with a low sampling ratio of 1/25. We can see that when the sampling ratio
is high, all these methods can produce a good result except that the SpaRCS method detects  incomplete silhouettes of video foregrounds.
When the sampling ratio is low,  the SpaRCS method totally fails and the SpLR method can reconstruct the video backgrounds but its detected video foregrounds are blurred and incomplete. However, the proposed H-TenPCA and PG-TenPCA models can well reconstruct a  sharp video and detect a relatively complete video foreground on each example.

\begin{table}[htp]
\centering
\scriptsize
\caption{The averaged results of different methods on various real videos.}
\begin{tabular}{cccccc}
\toprule[1pt]
SR & \multicolumn{1}{c}{Indices} &\multicolumn{1}{c}{SpaRCS}  & \multicolumn{1}{c}{SpLR} & \multicolumn{1}{c}{H-TenRPCA} & \multicolumn{1}{c}{PG-TenRPCA} \\ \hline
\multirow{3}{*}{1/5} & PSNR  &  27.50   &  38.26 &  \textbf{42.42}   & 41.13   \\
                      & SSIM  &  0.7812   &  0.9488 &   \textbf{0.9788}   & 0.9756 \\
                      & F-measure  & 0.1946 &    0.6269   &  0.6799    &\textbf{0.6803} \\ \hline
\multirow{3}{*}{1/10} & PSNR  &  23.48  & 32.34&  \textbf{39.41}   &38.13   \\
                      & SSIM  & 0.6206    &  0.8430  &   \textbf{ 0.9629}   & 0.9599\\
                      & F-measure  &  0.0424&   0.5368   &  0.6756    &\textbf{0.6760} \\ \hline
\multirow{3}{*}{1/20} & PSNR  & 19.53  &  26.84 & 35.01   & \textbf{ 35.51}    \\
                      & SSIM  &  0.4665    &   0.7049  &   0.9153   & \textbf{  0.9396} \\
                      & F-measure  &  0.0158 & 0.3105  & 0.6393    &\textbf{0.6536 } \\ \hline
\multirow{3}{*}{1/25} & PSNR  & 18.54  & 24.48  &  31.82  &\textbf{34.46}    \\
                      & SSIM  & 0.4203    & 0.6245 &    0.8575    & \textbf{ 0.9301} \\
                      & F-measure  &   0.0155&    0.2514   &  0.6213   &\textbf{ 0.6475} \\ \hline
\multirow{3}{*}{1/30} & PSNR  & 17.87  & 24.36 &  31.15   & \textbf{33.86 }  \\
                      & SSIM  & 0.3943     &0.6026 &    0.8239   & \textbf{0.9245  } \\
                      & F-measure  &  0.0150&  0.2369   &0.6202    &\textbf{ 0.6389 } \\
\bottomrule[1pt]
\end{tabular}
\label{realTab3}
\end{table}
\begin{table}[htp]
\centering
\scriptsize\vspace{-3mm}
\caption{Comparison with the online method ReProCS. ``syn" indicates the synthetic video. videos (f), (h), and (p) are shown
in Fig.~\ref{realData}.}
\begin{tabular}{cccccccc}
\toprule[1pt]
Video & Indices &  ReProCS & H-TenRPCA & PG-TenRPCA  \\ \hline
    \multicolumn{5}{c}{Sampling Ratio = 0.75}
\\ \hline
\multirow{3}{*}{syn}   & PSNR &  25.56    &  22.34& \textbf{41.6157} \\
                       & SSIM &  0.9612      &  0.5858    & \textbf{0.9890} \\
                       & F-measure &  \textbf{0.8767}    &  0.8664 &  0.8652   \\ \hline
\multirow{3}{*}{(f)}   & PSNR &  15.23    & 17.62  & \textbf{42.29}\\
                       & SSIM &  0.8160    &  0.4119 & \textbf{0.9891} \\
                       & F-measure &  0.8098   & \textbf{0.8218}  & 0.8120    \\ \hline
\multirow{3}{*}{(h)}   & PSNR &  16.93   & 23.84 & \textbf{42.71}\\
                       & SSIM &  0.8780   & 0.5312 & \textbf{0.9832} \\
                       & F-measure &  0.7136     & \textbf{0.8217}   & 0.7324   \\ \hline
\multirow{3}{*}{(p)}   & PSNR & 14.45    & 16.48 & \textbf{45.09}\\
                       & SSIM &  0.6838    &0.3295 & \textbf{0.9905} \\
                       & F-measure & 0.6845     & \textbf{0.8524}  & 0.8383    \\ \hline
  \multicolumn{5}{c}{Sampling Ratio = 0.5}
\\ \hline
\multirow{3}{*}{syn}   & PSNR &  25.32    & 19.12& \textbf{33.7437} \\
                       & SSIM &  0.9509    &    0.4322    & \textbf{0.9541}  \\
                       & F-measure &  \textbf{0.8730}     & 0.8569 & 0.8540   \\ \hline
\multirow{3}{*}{(f)}  & PSNR &  15.13   & 14.16 & \textbf{36.44} \\
                       & SSIM &   0.7948    &0.2739 & \textbf{0.9668} \\
                       & F-measure &  0.8021   & \textbf{0.8184}   & 0.8141    \\  \hline
\multirow{3}{*}{(h)}   & PSNR &  16.95    & 20.48 & \textbf{36.02} \\
                       & SSIM &  0.8760    & 0.3931& \textbf{0.9375} \\
                       & F-measure &  0.7129     & \textbf{0.8057}   & 0.7437   \\ \hline
\multirow{3}{*}{(p)}   & PSNR &  14.52   & 13.12  & \textbf{38.98}\\
                       & SSIM &  0.6146    & 0.2225   & \textbf{0.9688} \\
                       & F-measure &  0.6876     & \textbf{0.8395}  & 0.8289   \\ \hline
\multicolumn{5}{c}{Sampling Ratio = 0.25}
\\ \hline
\multirow{3}{*}{syn}   & PSNR &  \textbf{24.94}     & 17.22& 24.42 \\
                       & SSIM &  \textbf{0.9360}     & 0.3281 & 0.7054 \\
                       & F-measure &  \textbf{0.8617}    & 0.8548 & 0.8434  \\ \hline
\multirow{3}{*}{(f)}  & PSNR &  14.99   &  12.31 & \textbf{21.37} \\
                       & SSIM &  \textbf{0.7524}   & 0.1874 &  0.6263 \\
                       & F-measure &  0.7537     & 0.7820  & \textbf{0.7890} \\  \hline
\multirow{3}{*}{(h)}  & PSNR &  17.02   & 18.59 & \textbf{26.68} \\
                       & SSIM &  \textbf{0.8756}    &  0.3225 & 0.7092 \\
                       & F-measure &  0.7090      & \textbf{0.8110}
& 0.8073   \\ \hline
\multirow{3}{*}{(p)}  & PSNR &  13.69    &11.16& \textbf{20.67} \\
                       & SSIM &  0.4529    & 0.1548 & \textbf{0.5502} \\
                       & F-measure &  0.5284      & \textbf{0.8368} & 0.8151
\\
\bottomrule[1pt]
\end{tabular}
\label{onlineCmp}\vspace{-4mm}
\end{table}

\subsubsection{Comparison with the Online Method ReProCS}
We also compared our methods with the ReProCS method on four videos (synthetic video, Pedestrians, Cubicle, and Office).
For each video, we can find a frame sequence of video background to train an initialized background for the ReProCS method. The compressive operator is chosen as the randomly permuted Walsh-Hardmard transform in the frame-wise manner, but set as the same for each frame
because of the constraint of the ReProCS method on compressive operator, i.e., $\mathcal{A}_d=\mathbf{\widetilde{D}} \cdot \mathbf{\widetilde{H}} \cdot \mathbf{\widetilde{P}}~(d=1,2,\cdots,D)$. The sampling ratios in this group of experiments are set as 0.75, 0.5, and 0.25, respectively.

\begin{figure}[htp]
\centering\vspace{-2mm}
\includegraphics[width=3.1in,height=2.45in]{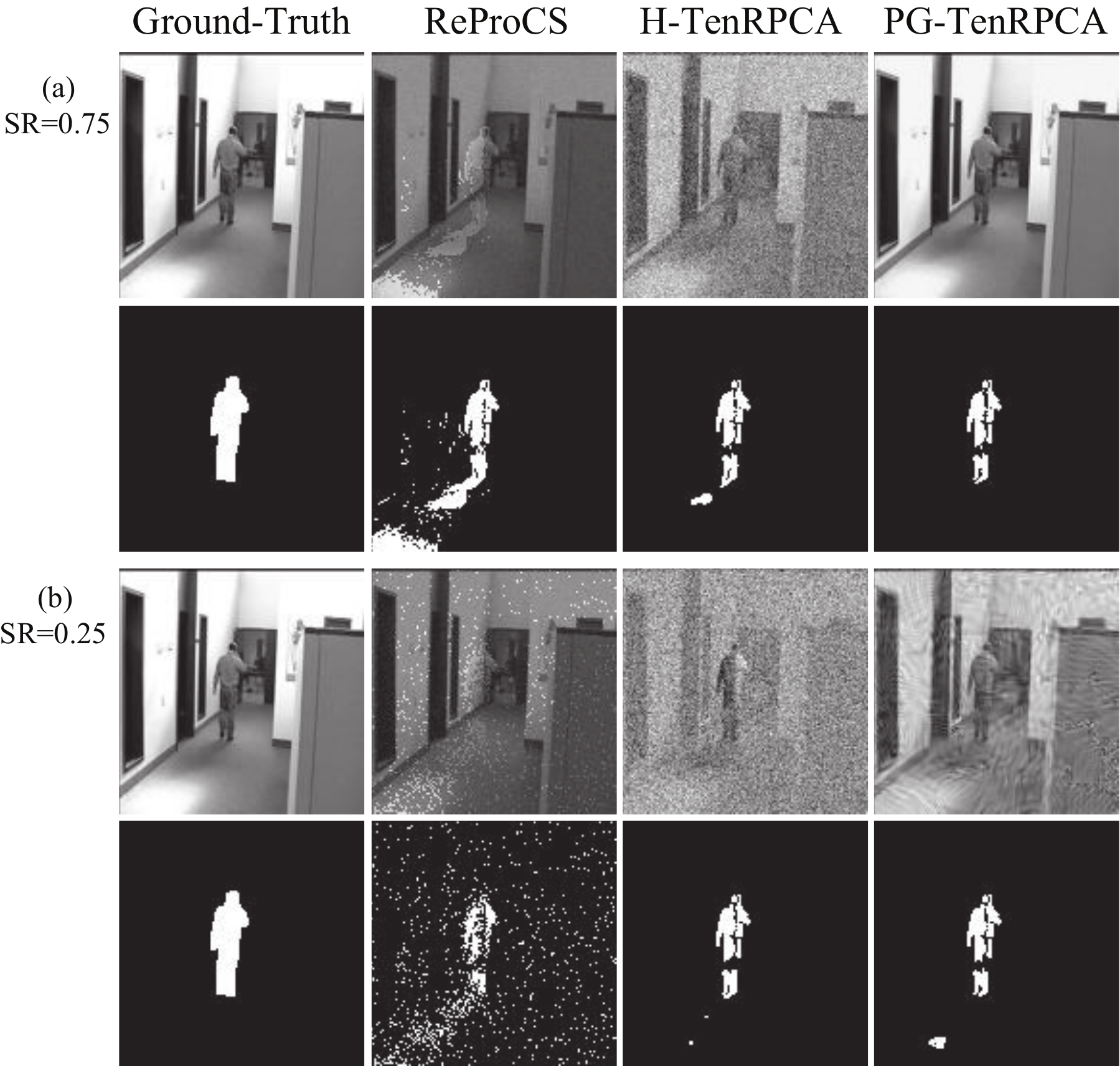}
\caption{Comparison of our methods H-TenRPCA and PG-TenRPCA with the online method ReProCS in reconstruction and separation performance.}
\label{onlineImag}\vspace{-2mm}
\end{figure}

We exhibit the reconstruction and separation results in Table~\ref{onlineCmp}. From this table, we can find that our proposed methods almost outperform the ReProCS method in terms of F-measure index for video separation (foreground detection). This good detection performance on video foreground can be attributed to the favor of spatio-temporal continuity from 3D total variation. Moreover, for video reconstruction, the PG-TenRPCA method is superior over methods H-TenRPCA and ReProCS in terms of PSNR and SSIM indices. Additionally, when the sampling ratio is very low, on some videos, e.g., the synthetic video and video (p), the ReProCS method can reconstruct a better video than the H-TenPRCA method in terms of SSIM and PSNR indices. This is because the pre-trained video background provides sufficient information for the ReProCS method; however, for our proposed methods, the pre-training procedure is not required. These findings can be further supported in Fig.~\ref{onlineImag}, where we exhibit the reconstruction and separation results of one video with a high sampling ratio of 0.75 and a low sampling ratio of 0.25.

\subsubsection{Computational Speeds}
We compare the running time of different models on the resized video ``ShoppingMall'' of the size $64 \times 64\times 128$.
It is noted that this kind of comparison in terms of the running time is only illustrative. The running time (in seconds) for the BSCM task by SpaRCS, SpLR, ReProCS, H-TenRPCA, and PG-TenRPCA are 14.4606, 2147.6, 995.9870, 93.3779, and 1015.5 respectively.
It can be observed that the SpaRCS method is the fastest method among all the compared methods. But it cannot achieve comparable video reconstruction and separation performance  compared to the other methods. The ReProCS and SpLR methods require solving an expensive linear system, which is highly computational expensive. For our proposed basic model H-TenRPCA, as the compressive operator used in experiments is orthogonal in columns, then each sub-problem has the closed-form solution, which makes H-TenRPCA relatively fast. It is not hard to see that the search of similar 3D patches and the joint Tucker decomposition in PG-TenRPCA both consume expensive computational cost, which makes it relatively slow. However, as shown in Fig.~5, for each cluster composed of similar 3D patches, all related  computations can be performed in a parallel way. Then, the computational cost will be greatly reduced if more processors are provided. The parallelization of our optimization algorithm for PG-TenRPCA deserves us to investigate in the future work.

\subsection{Effect of Compressive Operators}
In this subsection, we will report the reconstruction and separation performance of our proposed models based on the measurements of videos captured with different compressive operators.
The compressive operator is chosen as the randomly permuted Walsh-Hardamard transform in the holistic and frame-wise manner (WHT-h and WHT-f), and the randomly permuted noiselet transform in the holistic and frame-wise manner (Noiselet-h and Noiselet-f), respectively.

In Table~\ref{simuOperator} and \ref{realOperator}, we show the quantitative results of the proposed models with different sampling ratios on the synthetic and ``ShoppingMall" video, respectively. From these two tables, we can observe that when the sampling ratio goes down, the  results based on these four compressive operators deteriorate in terms of PSNR, SSIM and F-measure; the results based on WHT-h (WHT-f) is comparable to those based on Noiselet-h~(Noiselet-f); and the results based on the compressive operator in the holistic manner (WHT-h and Noiselet-h) are slightly better than
those based on the frame-wise compressive operator. It is worthy to point out that our proposed models can be incorporated with any compressive operator in the same framework.

\begin{table}[htp]
\setlength{\tabcolsep}{1.5pt}
\linespread{1.3}
\centering
\tiny
\caption{Effect of different compressive operators on the synthetic video.}
\begin{tabular}{ccccccccc}
\toprule[1pt]
\multirow{2}{*}{SR} &\multicolumn{2}{c}{WHT-f}  & \multicolumn{2}{c}{WHT-h} & \multicolumn{2}{c}{Noiselet-f} & \multicolumn{2}{c}{Noiselet-h} \\
  \cline{2-9}
   & H-TenRPCA & PG-TenRPCA & H-TenRPCA & PG-TenRPCA&   H-TenRPCA & PG-TenRPCA &  H-TenRPCA & PG-TenRPCA \\   \hline
\multirow{3}{*}{1/5} &  42.33  &   40.15  &     42.48 &    43.12   &     42.36   &  43.39   &    42.46  &  43.78\\
                &0.9918 &   0.9894   &   0.9922  &  0.9945   &    0.9919  &  0.9947  &     0.9921  & 0.9953 \\
               & 0.8624  &  0.8598   &   0.8616  &  0.8588   &    0.8605  &  0.8582  &     0.8614  & 0.8585 \\ \hline
\multirow{3}{*}{1/10} &34.95 &     34.38   &   34.96   &   35.86    &   35.18 &    35.54    &   35.33    &35.67 \\
                      &0.9695&     0.9673  &   0.9704   &  0.9767    &  0.9705 &    0.9746  &   0.9722   & 0.9756 \\
                      &0.8616 &    0.8566  &   0.8604    & 0.8638    &  0.8608  &   0.8603   &  0.8615   & 0.8614   \\ \hline
\multirow{3}{*}{1/20}    & 30.64   &    30.40    &  30.71    &  31.32      &     30.88     &  31.35    &   30.81    &  31.37 \\
                          &0.9275  &    0.9299   &  0.9307   &   0.9459     &    0.9274    & 0.9457    &   0.9311    & 0.9461 \\
                          &0.8337   &   0.8342    & 0.8277    &  0.8318     &    0.8345     & 0.8374    &   0.8290    & 0.8357  \\ \hline
\multirow{3}{*}{1/25}   & 28.80  &    29.61   &  28.87  &    30.31     &    28.96   &  30.34    &   28.89   &  30.21\\
                         & 0.8901  &   0.9236   & 0.8966  &   0.9349      &  0.8818   & 0.9350  &    0.8962  &   0.9337\\
                         & 0.8182  &   0.8240   & 0.8132   &  0.8221       & 0.8202   & 0.8272   &   0.8137  &   0.8219 \\ \hline
\multirow{3}{*}{1/30}   &27.36  &    28.88  &   27.72    &   29.46     &   27.05    & 29.52  &      28.03   &  29.47\\
                        &0.8492  &   0.9133  &  0.8635  &    0.9229    &   0.8189  &  0.9237   &    0.8710   &  0.9227\\
                       & 0.8086  &   0.8148  &  0.8006   &   0.8114    &   0.8100  &  0.8168    &   0.8024   &  0.8171 \\
\bottomrule[1pt]
\end{tabular}
\label{simuOperator}
\end{table}
\begin{table}[htp]
\setlength{\tabcolsep}{1.5pt}
\linespread{1.3}
\centering
\tiny
\caption{Effect of different compressive operators on the ``ShoppingMall" video.}
\begin{tabular}{ccccccccc}
\toprule[1pt]
\multirow{2}{*}{SR} &\multicolumn{2}{c}{WHT-f}  & \multicolumn{2}{c}{WHT-h} & \multicolumn{2}{c}{Noiselet-f} & \multicolumn{2}{c}{Noiselet-h} \\
  \cline{2-9}
   & H-TenRPCA & PG-TenRPCA & H-TenRPCA & PG-TenRPCA&   H-TenRPCA & PG-TenRPCA &  H-TenRPCA & PG-TenRPCA \\   \hline
\multirow{3}{*}{1/5} & 41.01 & 40.25  & 41.07 & 40.40& 40.94 & 40.43   & 41.07  & 40.53\\
                     & 0.9768 &0.9736 & 0.9772& 0.9745 &0.9767 & 0.9745  & 0.9771 & 0.9761  \\
                     & 0.6714  &0.6672  & 0.6697 & 0.6671 & 0.6715    & 0.6701 & 0.6707  & 0.6680  \\ \hline
\multirow{3}{*}{1/10} &  36.94 &36.38  & 37.06& 36.45 &  36.93& 36.47 & 37.07 & 36.48  \\
                      & 0.9569 &  0.9535  & 0.9583 &   0.9549& 0.9576& 0.9549  & 0.9588 & 0.9554 \\
                     & 0.6635  & 0.6568   &0.6626   &   0.6591  &0.6591& 0.6565  &  0.6620 & 0.6609 \\ \hline
\multirow{3}{*}{1/20} &31.07  & 32.48  &  31.34  & 32.52& 31.29&  32.57  & 31.64  &  32.54\\
                      & 0.8715   &  0.9197  &   0.8779& 0.9197&  0.8726  & 0.9202   & 0.8861   &   0.9206\\
                     & 0.6186 & 0.6314  &0.6179  & 0.6345  & 0.6155 & 0.6309& 0.6221   &    0.6310 \\ \hline
\multirow{3}{*}{1/25} & 29.85 & 31.47  &  29.96 & 31.49 &28.87& 30.11  & 30.57 & 31.41  \\
                      & 0.8259  & 0.9058  &  0.8384 & 0.9081& 0.7740&  0.9047  &  0.8414 &  0.9060  \\
                     & 0.6008   &  0.6146 & 0.5941 &  0.6172 & 0.5922 &  0.6082  &  0.5964 & 0.6126 \\ \hline
\multirow{3}{*}{1/30} & 28.38 & 30.71 & 28.72 & 30.57&  26.73 & 29.49  &28.71 &  30.57 \\
                      & 0.7561 &  0.8948  & 0.7870 & 0.8940&0.6315 & 0.8911   & 0.7868 & 0.8916 \\
                     &0.5842  & 0.5964   & 0.5874  &  0.6073& 0.5677 &0.5852  &0.5764  &  0.5937   \\
\bottomrule[1pt]
\end{tabular}
\label{realOperator}
\end{table}
\section{Conclusion}
In this paper, we proposed a novel tensor-based robust PCA approach for background subtraction from compressive measurements, in which Tucker decomposition is utilized to model the spatio-temporal correlation of the background in video streams, and 3D-TV is employed to characterize the smoothness of video foreground. Furthermore, we proposed an improved tensor RPCA model that models the video background as  several tensors over groups of similar video patches, taking advantages of the strong correlations of these patches in each patch group. Extensive experiments on synthetic and real-world data sets are conducted to demonstrate the superiority of proposed approaches over the existing state-of-the-art approaches.

In the future work, we are interested in the following research directions.  First, model the layers of the foregrounds using mixture of Gaussian to enhance its encoding capability for complex configured foreground. Second, develop better model for the complex background, such as dynamic background with illumination change, smog or snow,~and so on. Third, incorporate the motion of cameras into our proposed models. Finally, develop online version of our approach to make it more effective, thus facilitating the further use for more practical scenarios.

%

\appendices
\section{}
This optimization problem can be approximately solved by the alternating direction method (ADM). Firstly, fixing the orthogonal factors $\mathbf{U}_{1p}$, $\mathbf{U}_{2p}$, $\mathbf{U}_3$, and $\mathbf{U}_{4p}$, we have
$||\mathcal{R}_{p}(\widetilde{\mathcal{X}_1}) - \mathcal{G}_p \times_1 \mathbf{U}_{1p} \times_2 \mathbf{U}_{2p} \times_3 \mathbf{U}_3 \times_4 \mathbf{U}_{4p} ||_{F}^2 = ||\mathcal{R}_{p}(\widetilde{\mathcal{X}_1}) \times_1 \mathbf{U}_{1p}^{T} \times_2 \mathbf{U}_{2p}^{T} \times_3 \mathbf{U}_{3}^{T} \times_4\mathbf{U}_{4p}^{T} - \mathcal{G}_p ||_{F}^{2}$. Hence, it follows that $$\mathcal{G}_p= \mathcal{R}_{p}(\widetilde{\mathcal{X}_1}) \times_1 \mathbf{U}_{1p}^{T} \times_2 \mathbf{U}_{2p}^{T} \times_3 \mathbf{U}_{3}^{T} \times_4 \mathbf{U}_{4p}^{T}.$$
Then, using the solution of $\mathcal{G}_p$ we further derive that
\begin{equation*}
\begin{split}
&\|\mathcal{R}_{p}(\widetilde{\mathcal{X}_1}) - \mathcal{G}_p \times_1 \mathbf{U}_{1p} \times_2 \mathbf{U}_{2p} \times_3 \mathbf{U}_3 \times_4 \mathbf{U}_{4p} \|_{F}^{2}= \\
&\|\mathcal{R}_{p}(\widetilde{\mathcal{X}_1})\|_{F}^{2} -2 \langle \mathcal{R}_{p}(\widetilde{\mathcal{X}_1}),\mathcal{G}_p \times_1 \mathbf{U}_{1p} \times_2 \mathbf{U}_{2p} \times_3 \mathbf{U}_3 \times_4 \mathbf{U}_{4p} \rangle  \\
& + ||\mathcal{G}_p||_{F}^{2} = \\
&||\mathcal{R}_{p}(\widetilde{\mathcal{X}_1})||_{F}^{2} - ||\mathcal{R}_{p}(\widetilde{\mathcal{X}_1}) \times_1 \mathbf{U}_{1p}^{T} \times_2 \mathbf{U}_{2p}^{T} \times_3 \mathbf{U}_{3}^{T} \times_4 \mathbf{U}_{4p}^{T} ||_{F}^{2}.
\end{split}
\end{equation*}
The factor matrix $\mathbf{U}_{1p}$ can be estimated by maximizing $||\mathcal{R}_{p}(\widetilde{\mathcal{X}_1}) \times_1 \mathbf{U}_{1p}^{T} \times_2 \mathbf{U}_{2p}^{T} \times_3 \mathbf{U}_{3}^{T} \times_4 \mathbf{U}_{4p}^{T} ||_{F}^{2}$ with respect to $\mathbf{U}_{1p}$. It then easily follows that $\mathbf{U}_{1p}=\text{SVD}\big(\mathcal{R}_p(\widetilde{\mathcal{X}_1})\times_2 \mathbf{U}_{2p}^{T} \times_3 \mathbf{U}_{3}^{T} \times_4 \mathbf{U}_{4p}^{T})_{(1)},r_1\big)$. Here, $\text{SVD}(\mathbf{A},r)$ indicates top $r$ singular vectors of matrix $\mathbf{A}$. Likewise, we can obtain the solutions for factor matrixes $\mathbf{U}_{2p}$ and $\mathbf{U}_{4p}$. Finally, the factor matrix $\mathbf{U}_3$ can be estimated by maximizing $\sum_p||\mathcal{R}_{p}(\widetilde{\mathcal{X}_1}) \times_1 \mathbf{U}_{1p}^{T} \times_2 \mathbf{U}_{2p}^{T} \times_3 \mathbf{U}_3 \times_4 \mathbf{U}_{4p}^{T} ||_{F}^{2}$ with respect to $\mathbf{U}_3$. It is easy to find that $\mathbf{U}_{3}=\text{eigs}\big(\sum_{p=1}^{P}\mathbf{Z}_p\mathbf{Z}_{p}^{T},r_3\big)$, where $\mathbf{Z}_p = \big(\mathcal{R}_p(\widetilde{\mathcal{X}_1}) \times_1 \mathbf{U}_{1p}^{T} \times_2 \mathbf{U}_{2p}^{T} \times_4 \mathbf{U}_{4p}^{T}\big)_{(3)}$ and $\text{eigs}(\mathbf{A},r)$ indicates top $r$ eigen vectors of matrix $\mathbf{A}$.

\section*{Acknowledgment}
The authors would like to thank the associate editor and the anonymous reviewers for their insightful comments,
which led to a significant improvement of this paper. The authors also would like to thank Dr. Waters, Dr. Deng, as well as Dr. Guo for sharing the codes of the SpaRCS method, the SpLR method, and the ReProCS method, respectively.

\ifCLASSOPTIONcaptionsoff
  \newpage
\fi



%

\begin{IEEEbiography}[{\includegraphics[width=1.02in,height=1.21in]{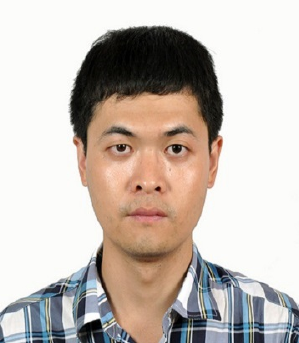}}]{Wenfei Cao}
received the Ph.D. degree in computer science from Xi'an Jiaotong University, China, in 2015. He worked as a visiting student in RIKEN BSI, Japan, from Nov. 2012 to Oct. 2013. He is currently an assistant professor at the Department of Applied Mathematics in ShaanXi Normal University. His current research interests include sparse optimization, randomized numerical linear algebra, and computer vision.
\end{IEEEbiography}
\begin{IEEEbiography}[{\includegraphics[width=0.98in,height=1.37in]{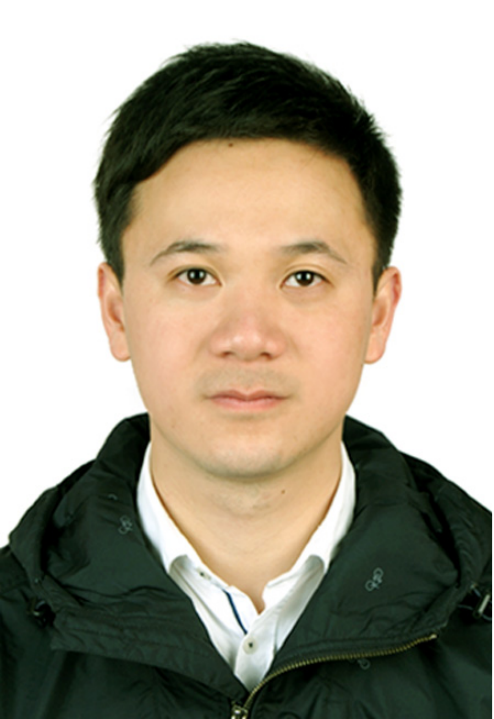}}]{Yao Wang}
received the Ph.D. degree in applied mathematics from Xi'an Jiaotong
University, China, in 2014. He worked as a visiting student in Georgia Institute of Technology form Oct. 2010 to Nov. 2011. He is currently an assistant professor at the Department of Statistics in Xi'an Jiaotong University. His current research interests include statistical signal
processing, high-dimensional statistical inference, large-scale video analysis, and computational biology.
\end{IEEEbiography}
\vspace{-8mm}
\begin{IEEEbiography}[{\includegraphics[width=1.0in,height=1.22in]{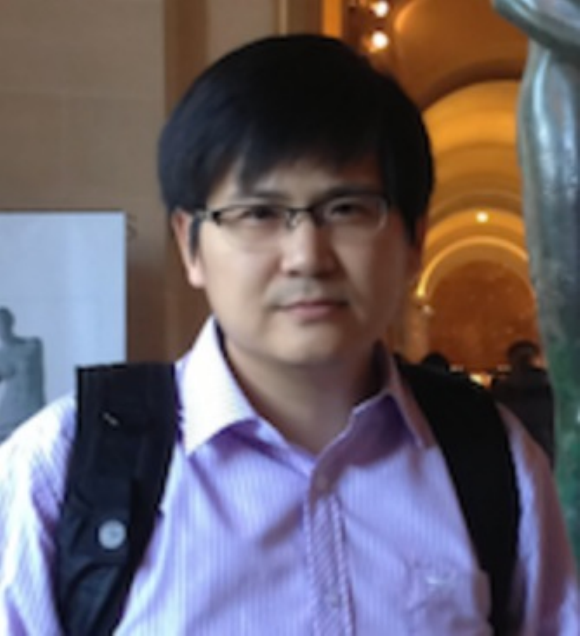}}]{Jian Sun}
received the Ph.D. degree in applied mathematics from Xi'an Jiaotong University, China, in 2009. He worked as a visiting student in Microsoft Research Asia from Nov. 2005 to Mar. 2008, a post-doctoral researcher in University of Central Florida from Aug. 2009 to Apr. 2010, and a post-doctoral researcher in willow project team of \'{E}cole Normale Sup\'{e}rieure de Paris and INRIA from Sept. 2012 to Aug. 2014.  He is currently an associate professor with the Institute for Information and System Sciences, Xi'an Jiaotong University. His current research interests include computer vision and image processing.
\end{IEEEbiography}
\vspace{-8mm}
\begin{IEEEbiography}[{\includegraphics[width=1.02in,height=1.22in]{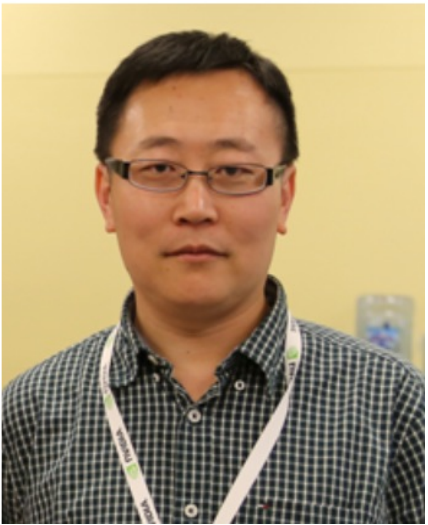}}]{Deyu Meng}
received the Ph.D. degree in computer science from Xi'an Jiaotong University, China, in 2008. He is currently an associate professor with the Institute for Information and System Sciences, Xi'an Jiaotong University. From 2012 to 2014, he took his two-year sabbatical leave in Carnegie Mellon University. His current research interests in￼￼￼clude self-paced learning, noise modeling, weekly supervised learning, and tensor sparsity.
\end{IEEEbiography}
\vspace{-8mm}
\begin{IEEEbiography}[{\includegraphics[width=1.02in,height=1.22in]{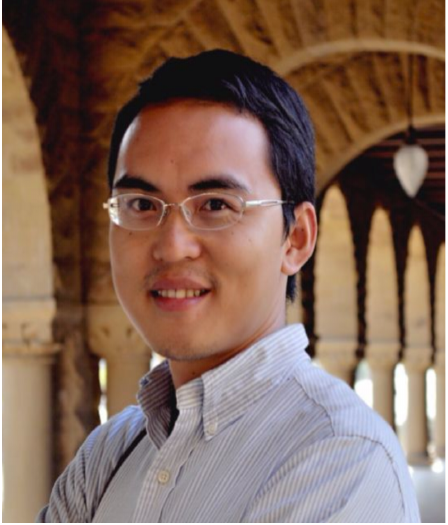}}]{Can Yang}
received the Ph.D. degree in electronic and computer engineering from
the Hong Kong University of Science and Technology in 2011. He worked as an associate researcher scientist at Yale University, New Haven, Connecticut. He is currently an assistant professor at the Department of Mathematics, Hong Kong Baptist University. His research interests include Biostatistics, machine learning, and pattern recognition.
\end{IEEEbiography}
\vspace{-8mm}
\begin{IEEEbiography}[{\includegraphics[width=1.03in,height=1.23in]{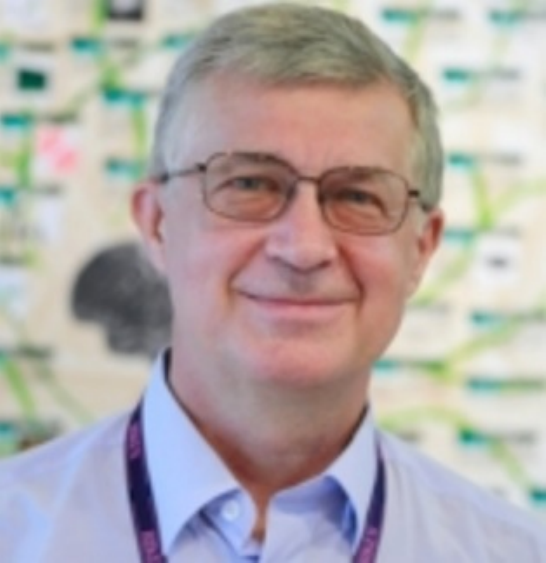}}]{Andrzej Cichocki } received the Ph.D. and Dr.Sc. (Habilitation) degrees, all in electrical engineering, from Warsaw University of Technology, Poland. He is currently the senior team leader of the Laboratory for Advanced Brain Signal Processing at RIKEN BSI, Japan. He is coauthor of more than 400 scientific papers and 4 monographs. He served as AE of IEEE Transactions on Signal Processing, Neural Networks and Learning Systems, Cybernetics. He is a fellow of the IEEE. 
\end{IEEEbiography}
\vspace{-8mm}
\begin{IEEEbiography}[{\includegraphics[width=1.02in,height=1.22in]{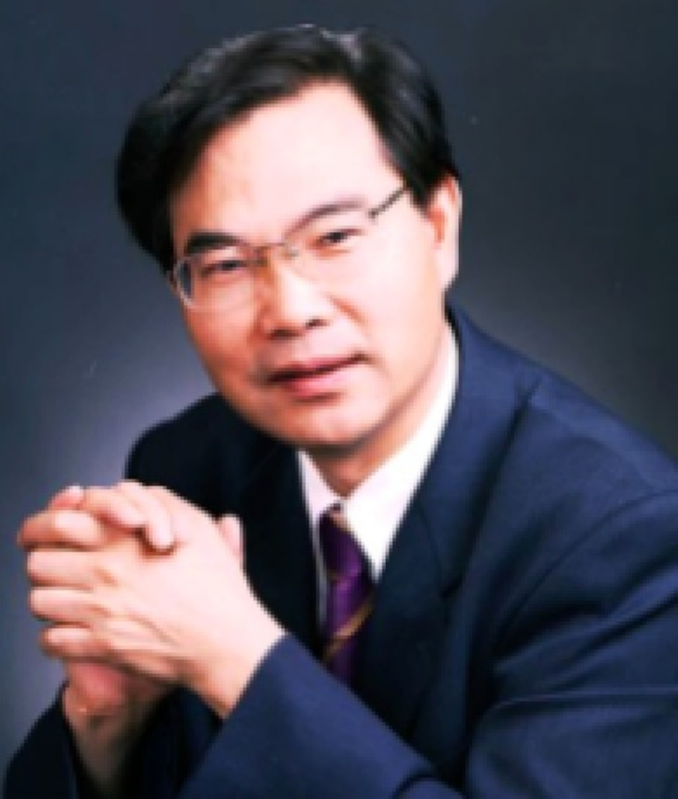}}]{Zongben Xu}
received his Ph.D. degree in mathematics from Xi'an Jiaotong University, China, in 1987. He now serves as the Chief Scientist of National Basic Research Program of China (973 Project), and Director of the Institute for Information and System Sciences of Xi'an Jiaotong University. He was elected as member of Chinese Academy of Science in 2011. His current research interests include intelligent information processing and applied mathematics.\end{IEEEbiography}
\end{document}